\documentclass{lcs2}
\usepackage{subcaption}
\usepackage{arydshln}
\usepackage{tabularx}
\usepackage{enumitem}
\usepackage{eso-pic}
\usepackage{makecell}
\definecolor{ForestGreen}{RGB}{34,139,34}

\usepackage{array} % For table wrapping
\usepackage{colortbl} % For coloring the table
\usepackage{pdflscape} % For landscape pages
\usepackage{booktabs}
\usepackage{multirow}
\papergithub{https://github.com/C0mRD/Counter_with_evidence}  %optional
\papertitle{\textit{Counter with Evidence}! \\ A Multi-Agent Memory Efficient Reasoning Framework for Hate Category Informed Counterspeech Generation} %change this
\papershorttitle{Accepted at EMNLP 2026 Main Conference}  %optional %change this
\papershortauthors{S. Nath, A. Kumar, and T. Chakraborty}

\paperauthors{%
  Sujoy Nath \textsuperscript{\tiny 1}\equalcontrib \quad
  Aswini Kumar \textsuperscript{\tiny 1}\equalcontrib \quad
  Tanmoy Chakraborty \textsuperscript{\tiny 1}%
}

\paperaffil{%
  \textsuperscript{1} Indian Institute of Technology Delhi, India\\
  * Equal contribution
}

\paperemails{%
  {\ttfamily\{sujoynathofficial,aswinikumarpadhi1995\}@gmail.com, tanchak@iitd.ac.in}
}

\paperkeywords{counterspeech generation, hate speech categorization, multi-agent reasoning} %optional %change this

\paperabstract{ 
Counterspeech effectively neutralizes the impact of online hate. Although prior work explores automated counterspeech generation, it largely emphasizes stylistic control while treating hate speech as homogeneous, overlooking that distinct forms of abuse require fundamentally different counterspeech strategies. To address this gap, we introduce \texttt{FIRE} (\textbf{F}actuality \textbf{I}nformed Multi-Agent \textbf{RE}asoning Framework) that first decomposes hate speech into one of the five distinct categories (misinformation, stereotype, conspiracy, dehumanizing, non-factual), and then maps it to a targeted counterspeech style. To facilitate \texttt{FIRE}, we curate \texttt{FactualCS}, a novel dataset of $4,784$ instances that provides the annotations regarding hate categories, reasoning traces, and evidence mappings, which are critical elements for grounded generation that are missing in prior work. A comprehensive evaluation across $28$ baseline configurations demonstrates that \texttt{FIRE} significantly surpasses existing methods, despite using compact agents ($<$2B). \texttt{FIRE} achieves a $\sim$ $12$\% and $\sim$ $11$\% improvements in factual and category-specific accuracy respectively, while simultaneously reducing toxicity by $\sim$ $11$\% relative to the strongest baselines. Further human evaluation confirms that responses generated by \texttt{FIRE} are significantly preferred over the strongest baselines, underscoring its effectiveness for real-world deployment. These findings show that decomposing the underlying intent of hate speech is essential for generating safe, effective, and contextually precise counterspeech.\footnote{\textcolor{red}{Warning: The materials presented in this paper might be disturbing or offensive.}}
} %change this

\appendixtocon % Appendix Table of Contents ON
\appendixtocname{Structure of the Appendix}   % optional
\begin{document}

\makelabtitle

%%%%%%%%%%%%%%%%%%%%%%%%%%%%%%%%%%%%%%%%%%%%%%%%%

\section{Introduction}

\begin{figure}[t]
\centering
\includegraphics[width=.5\textwidth]{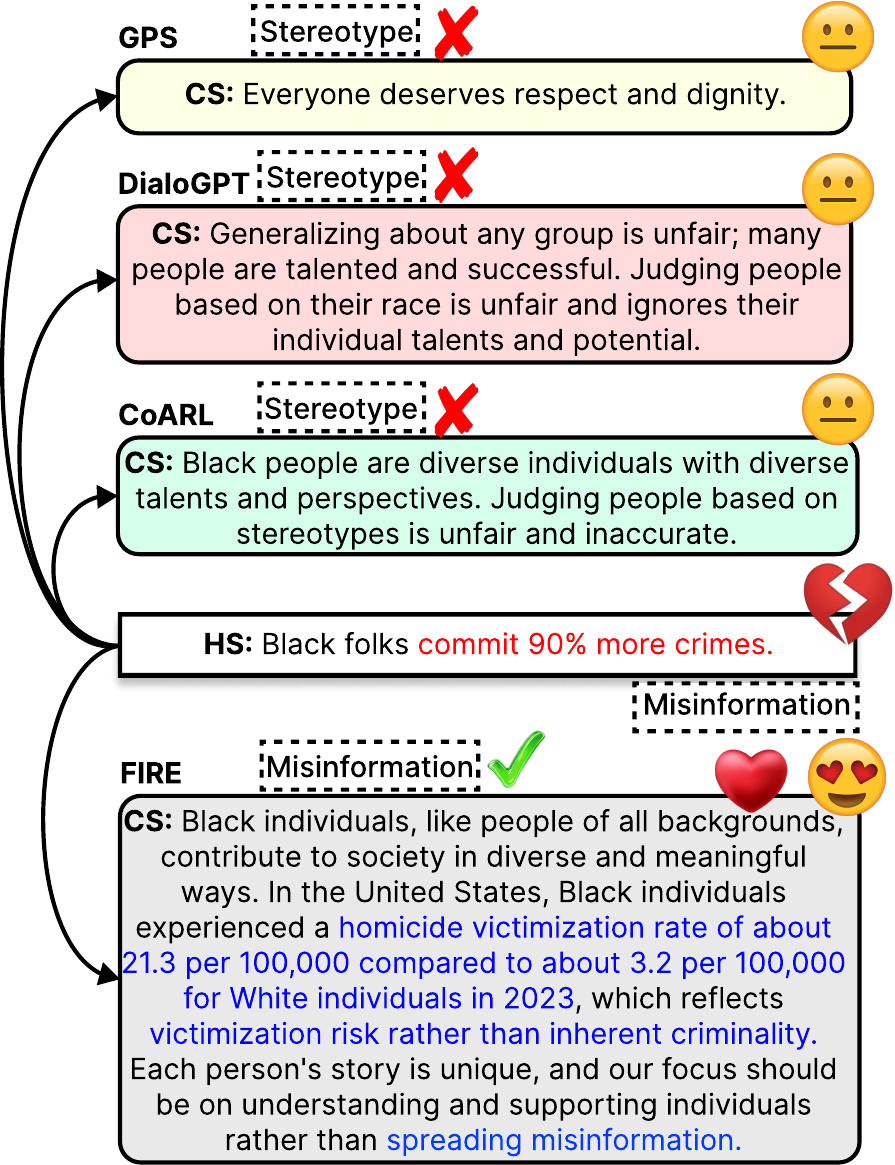}
\caption{A qualitative comparison demonstrating the efficacy of {\tt FIRE} against baselines; our framework generates superior quality response by accurately identifying the specific characteristics of the target hate speech.}
\label{fig:figure1}
\end{figure}

Social media platforms serve as the primary infrastructure for global communication \citep{bonaldi-etal-2024-nlp, gennaro2025counterspeech}. However, this connectivity is frequently weaponized to target vulnerable communities with Hate Speech (HS), causing severe psychological distress \citep{awal2021angrybert, davidson2017automated}. To address online hate, counterspeech (CS) comes out as an efficient approach, which aims to neutralize hostility through polite, factually grounded engagement rather than suppression \citep{benesch2016considerations, wright2017vectors}.

Given the sheer volume of online content, manual intervention is unscalable, making automated CS generation a critical research trajectory \citep{gupta-etal-2023-counterspeeches}. Early approaches treated this as a text-to-text translation task, mapping hateful inputs to safe replies \citep{chung2019conan, mathew2019thou}. Although subsequent research has moved beyond simple translation to control stylistic attributes of counterspeech like politeness or emotion \citep{ijcai2022p716}, these methods typically homogenize hate speech. By treating diverse abusive content as a uniform entity, they miss the critical requirement that different types of hate demand distinct logical refutations \cite{baider2023accountability}.

\paragraph{\textbf{Motivation and Crisis.}}
The primary limitation of current methodologies is their failure to account for the semantic diversity of hate speech, which is not uniform. Online hate manifests through distinct logical structures such as misinformation (verifiable false claims) \citep{wardle2024conceptual}, conspiracies (allegations of secret plots) \citep{jolley2020exposure}, stereotypes (oversimplified group generalizations) \citep{vargas2023socially}, dehumanization (denial of human dignity or animalistic metaphors) \citep{mendelsohn2020framework}, and non-factual assertions (subjective expressions of disgust or opinion) \citep{wardle2024conceptual}. Addressing these diverse forms of hate speech requires specific counter-strategies targeted to address that particular type of hate \citep{buerger2019counterspeech}. For instance, a verifiable false claim necessitates factual evidence, whereas a dehumanizing slur requires moral reframing. Current models, by failing to distinguish between these categories, often produce a generalized responses that fail to address the core premise of the abuse.

\paragraph{\bf Need of an Agentic Framework.} However, implementation of such specific strategies reveals the architectural limitations of monolithic Large Language Models (LLMs). A single LLM often struggles to simultaneously perform deep semantic reasoning (e.g., distinguishing \textit{misinformation} from a \textit{stereotype}) and factual verification while maintaining stylistic coherence \citep{liu2023pre}. Without explicit task decomposition, these models lack the capacity to plan their logical strategy, frequently hallucinating facts or reverting to safe, context-agnostic responses \citep{hengle2024intentconditioned}. To address this, we move to an \textit{agentic} paradigm. By employing specialized agents, we decouple the problem, the system first diagnoses the specific nature of the abuse before attempting to generate an appropriate counter-response. This mimics human cognitive decomposition, ensuring that responses are planned and factually grounded before they are articulated \citep{cima2025contextualized}. Crucially, training agents to reason rather than just mimic requires data that goes beyond the simple input-output pairs found in existing datasets \citep{saha2024crowdcounter, chung2019conan}. These resources lack the intermediate \textit{reasoning traces}, the explicit mapping between hate category, the search query required to debunk it, and the retrieved evidence, necessary to supervise an agent's logic.

\paragraph{\textbf{Our Major Contributions.}} 
To bridge these gaps, we introduce the Factuality Informed Multi-Agent Reasoning Framework (\texttt{FIRE}), which prioritizes the semantic decomposition of abuse. \texttt{FIRE} employs specialized agents ($<$2B), to decouple the reasoning phase from the generation phase. This allows the system to explicitly identify one of five hate categories (Misinformation, Stereotype, Conspiracy, Dehumanization, Non-factual). This modularity ensures the response strategy matches the specific abuse type. To support \texttt{FIRE}, we develop \texttt{FactualCS}, a dataset comprising 4,784 hate-counterspeech pairs across 14 target communities. Unlike prior resources, \texttt{FactualCS} includes dense annotations for hate categories, reasoning traces, and evidence mappings. This solves the data scarcity issue for grounded counterspeech, enabling models to learn \textit{why} a response is effective, not just \textit{what} to say. As demonstrated in Figure \ref{fig:figure1}, \texttt{FIRE} outperforms baselines like GPS \citep{zhu-bhat-2021-generate}, DialoGPT \citep{zhang2020dialogpt}, and CoARL \citep{hengle2024intentconditioned}, which often produce context-agnostic text. Extensive evaluation shows that \texttt{FIRE} achieves state-of-the-art performance, matching larger models like LLaMA-3.1-8B while significantly improving factuality and category-specific accuracy.

\section{Related Works}
The study of counterspeech generation has evolved through a synergy of dataset curation and architectural advancements. Early efforts relied on expert-crafted, static corpora like CONAN \cite{chung2019conan} and Multi-Target CONAN \cite{fanton2021human}, which provided high-quality but limited coverage. Recent benchmarks such as CrowdCounter \cite{saha2024crowdcounter}, IntentCONANv2 \cite{hengle2024intentconditioned}, and MultiCONAN \cite{padhi-etal-2025-counterspeech} introduce annotations for intent and strategy, that support the modeling of diverse response strategies. Methodologically, counterspeech generation has progressed from basic sequence-to-sequence translation tasks \cite{qian2019benchmark, mathew2019thou} toward frameworks prioritizing strategic control. Recognizing that a single generic response is often insufficient, researchers developed pipelines like GPS \cite{zhu-bhat-2021-generate} and CounterGeDi \cite{ijcai2022p716}, which steer generation toward specific tones such as politeness or detoxification. \textcolor{black}{Moving beyond mere text fluency, recent strategies actively dismantle biases rather than generically denouncing hate \cite{gupta-etal-2023-counterspeeches, fraser-etal-2023-makes, mun2023beyond}. However, monolithic LLMs' struggles with reasoning and hallucination have catalyzed a shift toward Agentic AI \cite{acharya2025agentic} to better preserve human-like authenticity and empathy \cite{mun2024counterspeakers}.} Unlike isolated prompting, multi-agent frameworks employ iterative reasoning and task decomposition to handle the evolving complexity of hate speech \cite{chen2025static}. By facilitating self-correction and integrating external knowledge via Retrieval Augmented Generation (RAG) \cite{lewis2021retrieval}, these systems significantly reduce hallucinations and ensure interventions are factually grounded \cite{anik2025multi, podolak2024llm}. Our work builds upon this frontier, leveraging agentic decomposition to enable complex, evidence-based social strategies that are difficult to achieve with single-model architectures \cite{song2025assessing}.

\begin{table}[!ht]
\centering
\small
\begin{tabularx}{.5\columnwidth}{l *{6}{>{\centering\arraybackslash}X}}
\toprule
\textbf{Target} & \textbf{SH} & \textbf{MH} & \textbf{CH} & \textbf{DH} & \textbf{NH} & \textbf{Total} \\
\midrule
Women & 178 & 12 & 3 & 166 & 285 & 644 \\
Men & 29 & 3 & 1 & 14 & 15 & 62 \\
LGBT+ & 117 & 34 & 2 & 110 & 137 & 400 \\
Muslim & 271 & 101 & 22 & 90 & 99 & 583\\
Jews & 167 & 35 & 96 & 109 & 36 & 443\\
IMGT & 278 & 275 & 15 & 103 & 166 & 837\\
PoC & 232 & 17 & 5 & 172 & 55 & 481\\
AP & 15 & 1 & 0 & 5 & 1 & 22\\
Disable & 41 & 1 & 0 & 73 & 5 & 120\\
Individual & 55 & 30 & 26 & 57 & 84 & 252\\
Political & 56 & 16 & 13 & 34 & 41 & 160\\
Occupation & 6 & 5 & 1 & 7 & 8 & 27\\
OR & 10 & 7 & 0 & 12 & 9 & 38\\
Others & 115 & 44 & 28 & 172 & 356 & 715 \\
\midrule
\textbf{Total} & \textbf{1570} & \textbf{581} & \textbf{212} & \textbf{1124} & \textbf{1297} & \textbf{4784} \\
\midrule
Train & 1269 & 471 & 181 & 919 & 1072 & 3912 \\
Val & 123 & 50 & 15 & 87 & 108 & 383\\
Test & 178 & 60 & 16 & 118 & 117 & 489\\
\bottomrule
\end{tabularx}
\caption{Distribution of Hate Speech Types Across Targeted Groups: Stereotype Hate (SH), Misinformation Hate (MH), Conspiracy Hate (CH), Dehumanization Hate (DH), Non-factual Hate (NH) \protect\footnotemark[3]}
\label{tab:hs_distribution}
\end{table}
\footnotetext[3]{OR: Other Religions, AP: Affected Person, PoC: Person of Color}
% \vspace{-1.5mm}
\section{The \texttt{FactualCS} Dataset}\label{sec:data} 
To be truly effective, counterspeech must dismantle the core premises of the hate speech it addresses. Since abusive content varies significantly, ranging from false claims to attacks on human dignity, the rebuttal must be equally specialized. A response designed to counter a stereotype, for instance, may be ineffective against a conspiracy theory. Unfortunately, existing resources have largely overlooked this nuance, typically focusing on simple input-output pairings or the stylistic tone of the reply, while neglecting the specific semantic category of the hate itself. This limitation leaves models blind to the underlying nature of the abuse they are trying to neutralize. Driven by this gap, we develop \texttt{FactualCS} \footnote[4]{Dataset available at: \url{https://huggingface.co/datasets/Aswini123/FactualCS}}, a specialized dataset designed to ground responses in the specific characteristics of the hate speech. 

Six annotators ($4$ male, $2$ female) voluntarily participated in the annotation process, which followed a multi-stage protocol designed to establish a shared understanding of the task before full-scale annotation. Prior to annotation, the annotators reviewed relevant resources on online harassment and counterspeech and discussed the hate-speech taxonomy and corresponding annotation criteria. They then completed a $500$-instance pilot phase, independently annotating hate-speech type, annotation rationale, target group, retrieval query, and supporting evidence, where applicable. Following the independent annotation, cases exhibiting disagreement were collaboratively examined to clarify challenging category boundaries, particularly between stereotypes and misinformation, dehumanizing language and other forms of non-factual hate, and cases requiring empirical verification. Insights from the pilot phase were subsequently formalized into a common set of operational guidelines specifying category definitions, evidence requirements, and decision rules for ambiguous or overlapping cases. These guidelines were then followed during the main annotation phase for the remaining instances, with hate-speech type, rationale, target group, query formulation, and evidence collection manually annotated. Further details of the complete annotation procedure, category-specific criteria, and decision rules are provided in Appendix \ref{procedure_annotatioin}. The main annotation phase achieved a mean pairwise Cohen's $\kappa$ of approximately $0.915$ across the six annotators for $5$-way categorization (Table \ref{tab:interm_annotation}). \texttt{FactualCS} comprises $4,784$ unique instances, each explicitly annotated with one of five functionally distinct categories: misinformation, stereotype, dehumanization, conspiracy theories, and non-factual claims (refer to \hyperref[tab:hs_taxonomy]{Table \ref{tab:hs_taxonomy}} and Appendix \ref{ap:taxonomy_distinctions} for an analysis of their functional boundaries). By explicitly mapping these categories, we provide the necessary context for models to learn not only \textbf{what} to say, but also \textbf{why} a particular response is appropriate for a specific form of abuse. This structured approach moves beyond simple text generation, offering a roadmap for logical refutation. Further details regarding the dataset statistics and the annotation guidelines are provided in \hyperref[tab:hs_distribution]{Table~\ref{tab:hs_distribution}} and Appendices \ref{data_stat}, and \ref{procedure_annotatioin}.

\section{Proposed Methodology}
\label{sec:proposed_methodology}
\begin{figure*}[t]
  %\begin{minipage}[c]
  %{0.69\textwidth}
    \centering
    \includegraphics[width=1\textwidth]{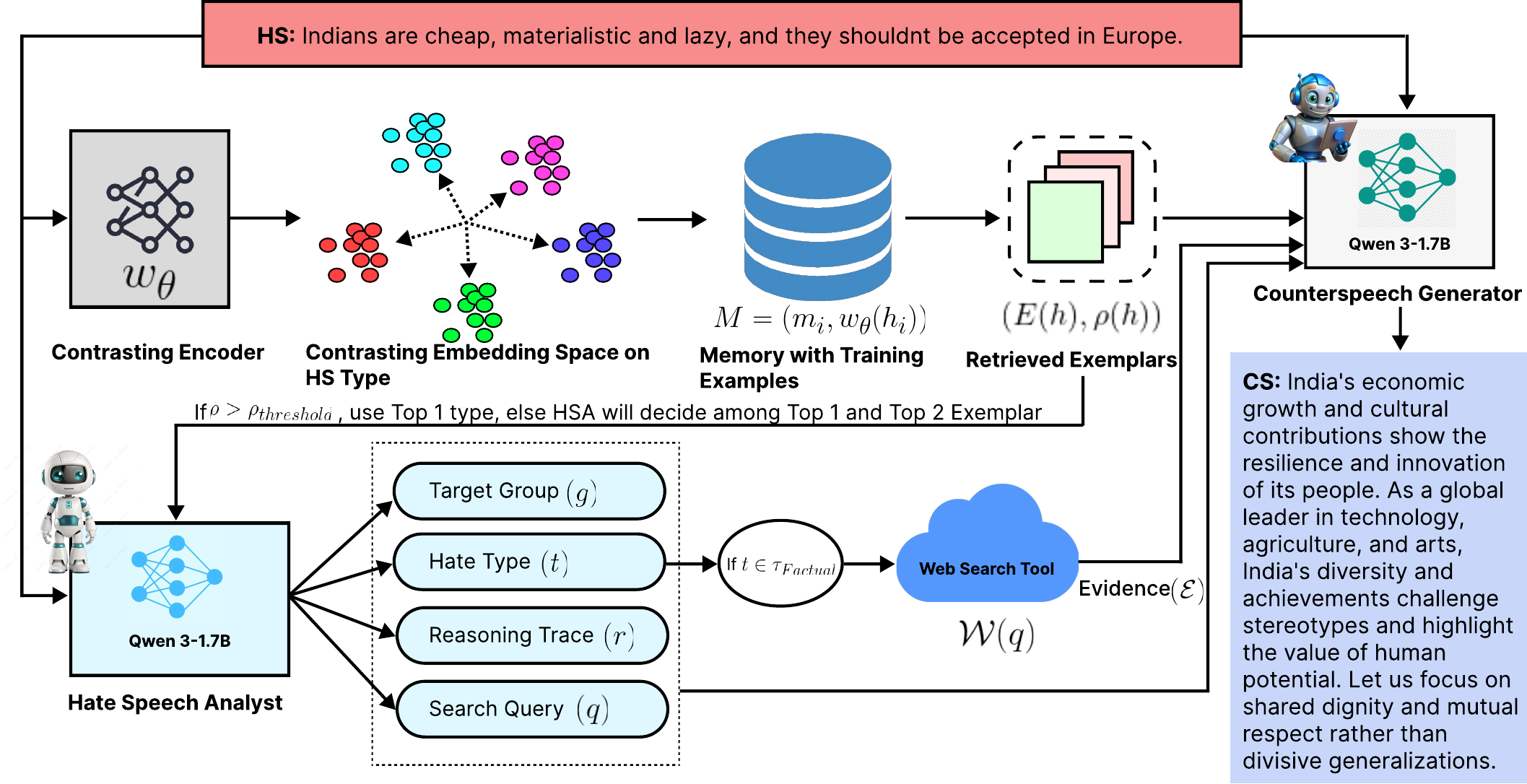}
  %\end{minipage}\hfill
  %\begin{minipage}[c]{0.3\textwidth}
\caption{\textbf{An overview of the proposed \texttt{FIRE} framework that follows the two-phase approach. }The Memory Module utilizes a trained contrasting encoder ($w_\theta$) to retrieve top-$k$ exemplars ($E$) and compute a purity score ($\rho$). In Phase 1 (HEAL), the Hate Speech Analyst (HSA) leverages these exemplars to predict structured latent variables ($g$, $t$, $r$, $q$). If the detected hate type contains factual claims ($t \in \mathcal{T}_{\text{factual}}$), a Web Search Tool is triggered to retrieve supporting evidence ($\mathcal{E}$). In Phase 2 (CARE), the Counterspeech Generator (CSG) synthesizes the final response by conditioning on the original input, the structured analysis from HSA, the retrieved evidence, and the stylistic patterns from the memory exemplars.} \label{fig:figure1_9}
 % \end{minipage}
  
\end{figure*}

% Overview of the proposed \texttt{FIRE} agentic framework follows the two-phase approach. First, the Memory Module utilizes a trained contrasting encoder ($w_\theta$) to retrieve top-$k$ exemplars ($E$) and compute a purity score ($\rho$) based on hate-type clustering. In Phase 1 (HEAL), the Hate Speech Analyst (HSA) leverages these exemplars to predict structured latent variables: target group ($g$), hate type ($t$), reasoning trace ($r$), and search query ($q$). If the detected hate type contains factual claims ($t \in \mathcal{T}_{\text{factual}}$), a Web Search Tool is triggered to retrieve supporting evidence ($\mathcal{E}$). In Phase 2 (CARE), the Counterspeech Generator (CSG) synthesizes the final response by conditioning on the original input, the structured analysis from HSA, the retrieved evidence, and the stylistic patterns from the memory exemplars.
% In this section, we detail the internal architecture of \texttt{FIRE}, our novel modular framework for generating counterspeech. We designed this system to overcome established limitations by focusing on two key mechanisms: (i) ensuring categorical precision, where the model tailors its response to the specific type of abuse found in the input, and (ii) enforcing quality alignment, which guides the generation to match the constructive, non-toxic nature of human-written response. An overview of this complete workflow is illustrated in Figure \ref{fig:figure1_9}.
In this section, we describe the design of our novel agentic framework \texttt{FIRE} for generating counterspeech. By concentrating on two crucial mechanisms, we designed \texttt{FIRE} to overcome known limitations: (i) ensuring categorical accuracy, where the model adapts its response to the particular type of abuse found in the input, and (ii) enforcing quality alignment, which directs the generation to match the constructive, non-toxic nature of human-written response. Figure \ref{fig:figure1_9} provides an overview of this entire workflow.

\subsection*{Problem Formulation}

We define the task of counterspeech generation as a conditional text generation problem. Let $\mathcal{D} = \{ d_1, d_2, .... dn\}$ represent our dataset, consisting of $n$ samples in which each data point is denoted as a tuple $(h_i, t_i, g_i, r_i, q_i, e_i, c_i)$. In this framework, $h_i \in \mathcal{H}$ serves as the sole input source (hate speech). The goal is to generate the target counterspeech $c_i \in \mathcal{C}$. However, to ensure the response is grounded and logically sound, the model must also generate a set of latent auxiliary variables: the hate category $t_i \in \mathcal{T}$, the target group $g_i \in \mathcal{G}$, the reasoning trace $r_i \in \mathcal{R}$, the search query $q_i \in \mathcal{Q}$, and the supporting evidence $e_i \in \mathcal{E}$. We denote this auxiliary context collectively as $\mathcal{Z} = \mathcal{T} \times \mathcal{G} \times \mathcal{R} \times \mathcal{Q} \times \mathcal{E}$. Consequently, our objective is to learn a generation function $\psi: \mathcal{H} \to \mathcal{Z} \times \mathcal{C}$, which maximizes the joint probability of generating both the intermediate context and the final response, formulated as $(z_i, c_i) \sim \psi(\cdot | h_i)$.

% Given a hate speech instance x, the goal of \textsc{hatebreak} is to generate a counterspeech y that is aligned with both the content of $x$ and its underlying factual mechanism. Each data point in our dataset is represented as $(x, g, t, r, q, w, y)$, where $g$ denotes the target group referenced in hate speech, $t \in \{1, ..., 5\}$ is the factuality based HS type (stereotype, misinformation, conspiracy, non-factual, dehumanizing), $r$ is the reasoning trace, $q$ a websearch query, $w$ the supporting evidence (empty for non-factual cases). We model counterspeech generation as learning a mapping $F: x \rightarrow y$, where the output $y = F(x)$ must satisfy two criteria: (i) it should reflect the hate type $t$ inferred from $x$, and (ii) it should incorporate factual grounding when the hate speech can be countered with proper evidence. Formally, the system decomposes the generation process into a sequence of latent decisions: $x \rightarrow (g, t, r, q) \rightarrow (E, w) \rightarrow y$, where $E$ denotes retrieved exemplers from a memory. The learning objective is to produce $y = \arg\max_{y'} P(y' \mid x, t, r, E, w)$, where $y'$ denotes possible counterspeech candidates.

We address this problem by decomposing counterspeech generation into two stages utilizing Qwen3-1.7B \cite{yang2025qwen3} in our agents, chosen for their superior reasoning capabilities \cite{joshi2025comprehensive}. In the first stage, the Hatespeech Analyst ($<$2b) leverages a memory module to identify the hate type, target, and reasoning, dynamically triggering a web search for evidence when factual claims are detected. In the second stage, the Counterspeech Generator ($<$2b) synthesizes the response with these outputs, including retrieved examples and evidence, to produce a response with a tone adapted to the specific hate category. Crucially, as the agents are not fine-tuned, \texttt{FIRE} relies on the memory module to derive specialization entirely from retrieved examples and structured guidance.
% We address this problem by decomposing the counterspeech generation task into two stages. In the first stage, the system focuses on understanding the hate itself. Hatespeech Analyst ($<2$b), supported by examples retrieved from our memory module, identifies the likely hate type, the target group, and produces a short reasoning behind it. When hate contains a factual claim, it also creates a simple search query so that the system can find supporting information. Using this, a websearch tool is called to gather a small amount of evidence from the web. In the second stage, Counterspeech Generator ($<2$b) generates the counterspeech by combining the original hate text with everything learned in the first stage, the predicted hate type, the reasoning, the retrieved examples, and any factual evidence. Each hate type follows its own style of response, so the generator adapts its tone and structure accordingly. Throughout the process, we use Qwen3.1-1.7B \cite{yang2025qwen3} model in our agents due to it's superior reasoning performance \cite{joshi2025comprehensive}. A key component is the memory module, as the agents are not finetuned; all specialization is based on the retrieved examples from the memory module and the structured guidance.

% \subsection*{Phase 1: \textbf{H}ate \textbf{E}xplanation with \textbf{A}nalysis and \textbf{L}ookup (HEAL)}

\subsection*{\textbf{Memory Module}} \texttt{FIRE} maintains a compact memory constructed from all annotated training instances. Each instance $i$ contributes a record $m_i = (h_i, t_i, g_i, r_i, q_i, e_i, c_i)$, where only the hate speech $h_i$ is used for learning representations. A lightweight encoder $w_\theta$ of size $22$M maps $h_i$ to an embedding $u_i = w_\theta(h_i)$, and the memory $M = {(m_i, u_i)}$ forms a fixed retrieval index. The encoder is trained to cluster instances sharing the same hate type while separating those from different types. Given a minibatch of embeddings ${u_i}$ and corresponding HS type labels ${t_i}$, we compute the pairwise similarity matrix with temperature $\tau$, $s_{ij} = cos(u_i, u_j)/\tau$. For each anchor $i$, positives are all other samples $j \neq i$ with matching hate type $t_j = t_i$. The supervised contrastive loss used to train the encoder is, \[\mathcal{L} = -\frac{1}{|\mathcal{I}|} \sum_{i \in \mathcal{I}} \frac{1}{|\mathcal{P}(i)|} \sum_{p \in \mathcal{P}(i)} \log \frac{\exp(s_{ip})}{\sum_{a \in \mathcal{A}(i)} \exp(s_{ia}) } \,\]
where $\mathcal{P}(i)$ is set of positive examples for $i$, $\mathcal{A}(i)$ contains all other samples in the batch except $i$, and $\mathcal{I}$ is the set of samples that have at least one positive example in the batch. Although $|\mathcal{P}(i)|$ varies across batches, normalization by $|\mathcal{P}(i)|$ ensures uniform contributions. This objective encourages embeddings of the same hate type to remain close while pushing apart embeddings of different types.

% At retrieval time, for a new input $h_i$, the encoder computes $u = w_\theta(h_i)$, and cosine similarities $s_i = cos(u, u_i)$ are evaluated against all memory entries. We compute a class-level score for each hate type $c$ by averaging the top-k similarities within that class: \[\mathcal{S}_c = 1/k \sum_{i\in Topk(c)} s_i \]
% The highest scoring classes $\hat{c}$ determine which exemplers to retrieve: $E = Topk\{m_i:t_i = \hat{c}\}$. These retrieved entries provide reasoning patterns that guide Hate Speech Analyst and Counter Speech Generator. Importantly, the encoder $w_\theta$ is the only trained module in \texttt{FIRE}, preserving modularity and avoiding any finetuning of small language models.

At retrieval time, for a new input $h$, the encoder computes $u = w_\theta(h)$, and cosine similarities are evaluated against all memory entries. For each hate type $t \in \mathcal{T}$, we define the index set $\mathcal{I}_t = \{i : t_i = t\}$ and compute similarities $s_i = \frac{u^\top u_i}{\|u\| \|u_i\|}$ for all $i \in \{1, \ldots, |M|\}$. The top-k most similar indices within hate type $t$ are obtained as $\text{TopK}_t(h) = \arg\max_{J \subset \mathcal{I}_t, |J|=k} \sum_{j \in J} s_j$, and we compute a type-level score by averaging these top-k similarities: $\mathcal{S}_t(h) = \frac{1}{k} \sum_{i \in \text{TopK}_t(h)} s_i$. The predicted hate type is $\hat{t} = \arg\max_{t \in \mathcal{T}} \mathcal{S}_t(h)$, which determines the exemplars to retrieve: $E(h) = \{m_i : i \in \text{TopK}_{\hat{t}}(h)\}$. These retrieved entries provide reasoning patterns that guide both the Hate Speech Analyst and Counter Speech Generator. Importantly, the encoder $w_\theta$ is the only trained module in \texttt{FIRE}, preserving modularity and avoiding any finetuning of language models.

\subsection*{Phase 1: \textbf{H}ate \textbf{E}xplanation with \textbf{A}nalysis and \textbf{L}ookup (HEAL)}
\paragraph{\textbf{Hate Speech Analyst (HSA).}}
Given hate speech $h$, retrieved exemplars $E(h)$, and a purity score $\rho(h)$ measuring retrieval confidence, the analyst generates the target group $g$, hate type $t$, reasoning trace $r$, and search query $q$ according to the distribution $P_{\text{HSA}}(g, t, r, q \mid h, E, \rho)$. The purity score quantifies the homogeneity of the k-nearest neighbors in the memory. Let $\mathcal{N}_k(h) = \arg\max_{J \subset \{1,\ldots,|M|\}, |J|=k} \sum_{j \in J} s_j$ be the k-nearest neighbors of $h$. The purity is  computed as:
\begin{equation*}
\rho(h) = \frac{1}{k} \sum_{i \in \mathcal{N}_k(h)} \mathbb{F}[t_i = \hat{t}]
\end{equation*}
where $\mathbb{F}[\cdot]$ is the indicator function. When $\rho(h) \geq \rho_{\text{thresh}}$, the analyst performs maximum a posteriori estimation using exemplars from the single most confident hate type $\hat{t}$. When $\rho(h) < \rho_{\text{thresh}}$, indicating ambiguity, exemplars from both the top-two predicted hate types $\hat{t}_1$ and $\hat{t}_2$ are provided to the analyst, which must then explicitly choose between them and justify its decision. Formally, the structured output is sampled as $(g, t, r, q) \sim P_{\text{HSA}}(\cdot \mid h, E, \rho)$, where the prompt construction function encapsulates all contextual information. This design guides HSA through both exemplar-based reasoning and explicit uncertainty cues, significantly improving its capability despite its small size.
% The HSA is a small frozen language model (Qwen3-1.7B) that performs structured interpretation of the hate speech input. The analyst is able to make reliable predictions because it operates under two forms external guidance: (i) hate speech type aware retrieval from the memory module and (ii) example driven prompting. For a given HS $x$, the analyst recieves not only the hate text but also a set of retrieved exemplars $E$ retrieved from memory. These examplars contain the hate speech text, its annotated hate type, reasoning and for factual hate speech an extra search query. The analyst uses these examples as anchors for its own reasoning, imitating the structure and justification patterns of similar annotated cases. Formally, the analyst predicts four structured variables: $(g,t, r, q) = f_{HSA}(x, E)$. Instead of blindly trusting the nearest class, the analyst is passed a purity score: the proportion of topk neighbours belonging to the most similar hate type. \[purity(x) = 1/k \sum^k_{j=1}f\{t_{ij} = \hat{c}\}\]
% Where $t_{ij}$ is the hate type of neighbour $i_j$. When purity is low, HSA is given two candidate types top1 and top2 along with their respective exemplars and explicitly choose between them and justify its choice. This guides the analyst through both examples and uncertainty cues, significantly improving its capability despite its small size.

\paragraph{\textbf{Websearch Tool.}}
For hate types that can be better addressed with proper factual evidence, the search query $q$ generated by the analyst triggers an external websearch tool. We define the set of factual hate types as $\mathcal{T}_{\text{factual}} = \{\text{misinformation}, \text{stereotype}, \text{conspiracy}\}$. The websearch function $\mathcal{W}: \mathcal{Q} \to \mathcal{E} \cup \{\emptyset\}$ takes a query $q$ and returns evidence snippets. The evidence retrieval process is formalized as: \[e = \begin{cases}
\mathcal{W}(q) & \text{if } t \in \mathcal{T}_{\text{factual}} \\
\emptyset & \text{otherwise}
\end{cases} \] where $\mathcal{W}(q)$ executes the web query $q$ and returns factual evidence snippets $e \in \mathcal{E}$. For all non-factual forms of hate, we set $e = \emptyset$. \textcolor{black}{A manual audit of 100 retrieved instances confirms our websearch tool effectively suppresses misinformation, with 92\% of evidence originating from institutional or mainstream informational sources (Appendix \ref{app:web_search_audit}).}

\subsection*{Phase 2: \textbf{C}ategory-aware \textbf{A}gentic \textbf{R}esponse \textbf{E}ngine (CARE)} 
\paragraph{Counterspeech Generator (CSG).}
The counterspeech generator produces the final counterspeech $c$ conditioned on all structured information extracted by the HSA, the retrieved exemplars, and the evidence. Let $\mathcal{Z} = (g, t, r, E, e)$ denote the complete conditioning context. We model the generator with conditional distribution:
\begin{equation*}
P_{\text{CSG}}(c \mid h, \mathcal{Z}) = \prod_{i=1}^{|c|} P_\phi(c_i \mid c_{<i}, h, \mathcal{Z})
\end{equation*}
where $c = (c_1, \ldots, c_{|c|})$ is the counterspeech token sequence and $c_{<i}$ represents all preceding tokens. The contextual variables in $\mathcal{Z}$ collectively shape the generation: hate type $t$ activates type-specific response strategies, exemplars $E$ supply stylistic patterns from similar cases, and evidence $e$ introduces factual grounding when required. As the model parameters remain frozen, all task specialization emerges from how the conditioning context $\mathcal{Z}$ modifies the generation distribution, enabling hate type aware and evidence based responses without finetuning. Due to resource and accessibility constraints, we focus on open-weight models for reproducible and cost-effective evaluation; however, our model-agnostic framework can be extended to proprietary models via API integration, with systematic comparison left for future work.

% \paragraph{\textbf{Counterspeech Generator}}
% We have choosen a small (Qwen3-1.7B) frozen language model that produces the final counterspeech conditioned on structured information extracted by the HSA and the retrieved exemplers. Let $\phi_{CSG}$ denote the underlying causal language model with parameters fixed. Formally, we model the generator as sampling $y \sim P_{\phi_{CSG}}(y|x, g, t, r, E, w)$. The contextual variables $g, t, r, E, w$ collectively shape the conditioning state of the model: t activates hate type specific counter speech strategy (see \hyperref[tab:hs_taxonomy]{Table~\ref{tab:hs_taxonomy}}), $E$ supplies stylistic examples of how similar cases were previously countered, and $w$ introduces verified information when factual grounding is required. As the model parameters are frozen, all task specialization emerges from how these variables jointly modify the conditional ditribution, allowing the small generator to produce hate type aware and evidence based counterspeech without any finetuning.
\begin{table*}[!ht]
\centering
% \small
\resizebox{\textwidth}{!}
{
\setlength{\tabcolsep}{0.5mm}
\fontsize{7pt}{10pt}\selectfont
\begin{tabular}
{llccccccccccccc}
\toprule
\textbf{Model} & \textbf{P/A} & \multicolumn{3}{c}{\textbf{ROUGE $\uparrow$}} & \textbf{BS $\uparrow$} & \textbf{M $\uparrow$} & \textbf{R $\downarrow$} & \textbf{CoSIM $\uparrow$} & \textbf{T $\downarrow$} & \textbf{Nov $\uparrow$} & \textbf{Div $\uparrow$} & \textbf{CatAcc $\uparrow$} & \textbf{FSc $\uparrow$} \\
\cmidrule(lr){3-5}
& & \textbf{R1} & \textbf{R2} & \textbf{RL} & & & & & & & & & \\
\midrule
GPS & -- & \(0.103\) & \(0.032\) & \(0.129\) & \(0.831\) & \(0.179\) & \(0.257\) & \(0.472\) & \(0.035\) & \(0.489\) & \(0.643\) & \(0.422\) & \(0.467\) \\
DialoGPT & -- & \(0.191\) & \(0.087\) & \(0.188\) & \(0.863\) & \(0.210\) & \(0.231\) & \(0.518\) & \(0.021\) & \(0.618\) & \(0.765\) & \(0.563\) & \(0.529\) \\
CoARL & -- & \(0.207\) & \(0.101\) & \(0.202\) & \(0.865\) & \(0.213\) & \(0.212\) & \(0.549\) & \(0.019\) & \(0.649\) & \(0.788\) & \(0.604\) & \(0.579\) \\
HiPPrO & -- & \(0.235\) & \(0.104\) & \(0.217\) & \(0.879\) & \(0.227\) & \(0.198\) & \(0.574\) & \underline{\(0.018\)} & \(0.651\) & \(0.809\) & \(0.622\) & \(0.605\) \\
\midrule
BART Large & ZS & \(0.063\) & \(0.018\) & \(0.072\) & \(0.823\) & \(0.083\) & \(0.274\) & \(0.318\) & \(0.041\) & \(0.487\) & \(0.742\) & \(0.108\) & \(0.198\) \\
GPT2 XL & ZS & \(0.057\) & \(0.016\) & \(0.028\) & \(0.788\) & \(0.064\) & \(0.346\) & \(0.147\) & \(0.061\) & \(0.414\) & \(0.687\) & \(0.138\) & \(0.182\) \\
Flan-t5-xl & ZS & \(0.104\) & \(0.042\) & \(0.123\) & \(0.842\) & \(0.092\) & \(0.247\) & \(0.382\) & \(0.029\) & \(0.524\) & \(0.731\) & \(0.273\) & \(0.332\) \\
Llama-3.1-8B-instruct & ZS & \(0.269\) & \(0.060\) & \(0.177\) & \(0.876\) & \(0.216\) & \(0.163\) & \(0.528\) & \(0.022\) & \(0.646\) & \(0.837\) & \(0.219\) & \(0.455\) \\
Mistral-7B-instruct & ZS & \(0.277\) & \(0.068\) & \(0.180\) & \(0.877\) & \(0.206\) & \(0.145\) & \(0.521\) & \(0.021\) & \(0.636\) & \(0.855\) & \(0.204\) & \(0.472\) \\
Qwen-2.5-7B-instruct & ZS & \(0.263\) & \(0.058\) & \(0.173\) & \(0.876\) & \(0.181\) & \(0.151\) & \(0.545\) & \(0.056\) & \(0.659\) & \(0.849\) & \(0.189\) & \(0.626\) \\
\midrule
BART Large & FS & \(0.141\) & \(0.029\) & \(0.102\) & \(0.835\) & \(0.123\) & \(0.243\) & \(0.392\) & \(0.032\) & \(0.529\) & \(0.763\) & \(0.174\) & \(0.262\) \\
GPT2 XL & FS & \(0.072\) & \(0.021\) & \(0.039\) & \(0.823\) & \(0.083\) & \(0.308\) & \(0.251\) & \(0.049\) & \(0.462\) & \(0.723\) & \(0.204\) & \(0.254\) \\
Flan-t5-xl & FS & \(0.173\) & \(0.061\) & \(0.182\) & \(0.859\) & \(0.142\) & \(0.219\) & \(0.474\) & \(0.024\) & \(0.561\) & \(0.764\) & \(0.346\) & \(0.413\) \\
Llama-3.1-8B-instruct & FS & \(0.265\) & \(0.058\) & \(0.186\) & \(0.879\) & \(0.173\) & \(0.150\) & \(0.535\) & \(0.022\) & \(0.667\) & \(0.860\) & \(0.324\) & \(0.520\) \\
Mistral-7B-instruct & FS & \(0.268\) & \(0.067\) & \(0.191\) & \(0.871\) & \(0.179\) & \(0.131\) & \(0.521\) & \(0.038\) & \(0.662\) & \(0.859\) & \(0.356\) & \(0.548\) \\
Qwen-2.5-7B-instruct & FS & \(0.255\) & \(0.056\) & \(0.183\) & \(0.879\) & \(0.170\) & \(0.165\) & \(0.535\) & \(0.061\) & \(0.641\) & \(0.835\) & \(0.338\) & \(0.681\) \\
\midrule
BART Large & RB & \(0.193\) & \(0.091\) & \(0.164\) & \(0.842\) & \(0.158\) & \(0.217\) & \(0.448\) & \(0.026\) & \(0.563\) & \(0.804\) & \(0.237\) & \(0.303\) \\
GPT2 XL & RB & \(0.079\) & \(0.028\) & \(0.044\) & \(0.841\) & \(0.093\) & \(0.282\) & \(0.314\) & \(0.041\) & \(0.503\) & \(0.742\) & \(0.259\) & \(0.321\) \\
Flan-t5-xl & RB & \(0.234\) & \(0.073\) & \(0.193\) & \(0.862\) & \(0.183\) & \(0.210\) & \(0.521\) & \(0.019\) & \(0.603\) & \(0.773\) & \(0.382\) & \(0.472\) \\
Llama-3.1-8B-instruct & RB & \(0.302\) & \(0.078\) & \(0.203\) & \(0.873\) & \(0.230\) & \(0.133\) & \(0.538\) & \(0.019\) & \(0.644\) & \(0.847\) & \(0.366\) & \(0.612\) \\
Mistral-7B-instruct & RB & \(0.298\) & \(0.080\) & \(0.202\) & \(0.872\) & \(0.221\) & \(0.133\) & \(0.523\) & \(0.020\) & \(0.654\) & \(0.857\) & \(0.413\) & \(0.709\) \\
Qwen-2.5-7B-instruct & RB & \(0.264\) & \(0.062\) & \(0.182\) & \(0.878\) & \(0.190\) & \(0.128\) & \(0.548\) & \(0.038\) & \(0.655\) & \(0.849\) & \(0.379\) & \(0.715\) \\
\midrule
BART Large & SFT & \(0.236\) & \(0.081\) & \(0.175\) & \(0.867\) & \(0.192\) & \(0.183\) & \(0.419\) & \(0.021\) & \(0.558\) & \(0.817\) & \(0.227\) & \(0.227\) \\
GPT2 XL & SFT & \(0.080\) & \(0.030\) & \(0.059\) & \(0.825\) & \(0.067\) & \(0.135\) & \(0.130\) & \(0.045\) & \(0.517\) & \(0.825\) & \(0.260\) & \(0.260\) \\
Flan-t5-xl & SFT & \(0.304\) & \(0.104\) & \underline{\(0.235\)} & \(0.876\) & \(0.238\) & \(0.203\) & \(0.548\) & \underline{\(0.018\)} & \underline{\(0.669\)} & \(0.797\) & \(0.508\) & \(0.508\) \\
Llama-3.1-8B-instruct & SFT & \(\textbf{0.316}\) & \(\textbf{0.109}\) & \(\textbf{0.239}\) & \underline{\(0.880\)} & \underline{\(0.239\)} & \(0.130\) & \underline{\(0.558\)} & \underline{\(0.018\)} & \(0.650\) & \(0.860\) & \(0.623\) & \(0.623\) \\
Mistral-7B-instruct & SFT & \underline{\(0.313\)} & \underline{\(0.107\)} & \(0.231\) & \(0.879\) & \(0.237\) & \(0.129\) & \(0.542\) & \(0.020\) & \(0.651\) & \(0.861\) & \(0.620\) & \(0.763\) \\
Qwen-2.5-7B-instruct & SFT & \(0.307\) & \(0.104\) & \(0.230\) & \(0.878\) & \(0.224\) & \underline{\(0.127\)} & \(0.556\) & \(0.019\) & \(0.665\) & \underline{\(0.862\)} & \underline{\(0.632\)} & \underline{\(0.810\)} \\
\midrule
\textbf{\texttt{FIRE}-(2x1.7B) (Ours)} & -- & \(0.294\) & \(0.071\) & \(0.197\) & \(\textbf{0.886}\) & \(\textbf{0.247}\) & \(\textbf{0.118}\) & \(\textbf{0.574}\) & \(\textbf{0.016}\) & \(\textbf{0.730}\) & \(\textbf{0.873}\) & \(\textbf{0.702}\) & \(\textbf{0.969}\) \\
\midrule
- without Websearch Tool & -- & \(0.277\) & \(0.060\) & \(0.179\) & \(0.879\) & \(0.224\) & \(0.133\) & \(0.535\) & \(0.017\) & \(0.708\) & \(0.866\) & \(0.664\) & \(0.832\) \\
- without Memory & -- & \(0.262\) & \(0.043\) & \(0.163\) & \(0.874\) & \(0.214\) & \(0.152\) & \(0.490\) & \(0.018\) & \(0.692\) & \(0.847\) & \(0.599\) & \(0.795\) \\
- without HSA & -- & \(0.231\) & \(0.034\) & \(0.148\) & \(0.868\) & \(0.189\) & \(0.162\) & \(0.410\) & \(0.020\) & \(0.656\) & \(0.837\) & \(0.507\) & \(0.62\) \\
\midrule
$\Delta_{FIRE (Ours) - BestBaseline}$ & & \textcolor{red}{\(\downarrow0.022\)} & \textcolor{red}{\(\downarrow0.038\)} & \textcolor{red}{\(\downarrow0.042\)} & \textcolor{ForestGreen}{\(\uparrow0.006\)} & \textcolor{ForestGreen}{\(\uparrow0.008\)} & \textcolor{ForestGreen}{\(\downarrow0.009\)} & \textcolor{ForestGreen}{\(\uparrow0.016\)} & \textcolor{ForestGreen}{\(\downarrow0.002\)} & \textcolor{ForestGreen}{\(\uparrow0.061\)} & \textcolor{ForestGreen}{\(\uparrow0.011\)} & \textcolor{ForestGreen}{\(\uparrow0.07\)} & \textcolor{ForestGreen}{\(\uparrow0.159\)}\\
\bottomrule
\end{tabular}
}
\caption{Comparing \texttt{FIRE} with baselines across various evaluation metrics. Here, ↑ (\textit{resp.} ↓) denotes that higher (\textit{resp.} lower) is better.  \textbf{Bold} (\textit{resp.} \underline{underline}) indicates the best (\textit{resp.} second-ranked) performance. Here P/A represents different strategies used: ZS for Zero Shot Prompting, FS for Few Shot Prompting, RB for Retrieval-Based and SFT represent the Supervised Finetuning.}
\label{tab:results}
\end{table*}

% Comparing \texttt{FIRE} with baselines across various evaluation metrics. Here, ↑ (\textit{resp.} ↓) denotes that higher (\textit{resp.} lower) is better.  \textbf{Bold} (\textit{resp.} \underline{underline}) indicates the best (\textit{resp.} second-ranked) performance. $*$ shows our model \textit{significantly} outperforms (\(p < 0.05\)) the best baseline Llama-3.1-8B-instruct finetuned model. Here P/A represents different strategies used: ZS for Zero Shot Prompting, FS for Few Shot Prompting, RB for Retrieval-Based and SFT represent the Supervised Finetuning.
% \vspace{-4mm}
\section{Experimental Setup}
\subsection{Baselines}
% To evaluate the effectiveness of our proposed approach, we compare against several baseline methods. We examine  GPS \cite{zhu2021generate}, which operates through three stages autoencoding, grammar-based filtering, and final response selection. We also fine-tune DialoGPT \cite{zhang2020dialogpt} to enhance its ability to generate responses that remain coherent within the given context. In addition, we integrate CoARL \cite{hengle-etal-2024-intent}, for conversational strategy based counterspeech generation and HiPPrO \cite{padhi-etal-2025-counterspeech} the state-of-the-art method that uses hierarchical prefix learning combined with preference optimization to generate multi-attribute conditioned counterspeech. Furthermore, we conduct experiments across multiple settings with six model architectures, including the encoder–decoder models $\textbf{BART}_{Large}$ \cite{lewis2020bart} and $\textbf{FLAN-T5}_{XL}$ \cite{chung2024scaling}, as well as the decoder-only models $\textbf{GPT2}_{XL}$, $\textbf{Llama-3.1-Instruct}_{8B}$ \cite{dubey2024llama}, $\textbf{Mistral Instruct}_{7B}$ \cite{jiang2023mistral7b}, and $\textbf{Qwen-2.5-Instruct}_{7B}$ \cite{qwen2025qwen25technicalreport}. In zero-shot (ZS) setting, we evaluate these models without any task-specific training. For few-shot (FS) evaluation, we provide five representative training examples as in-context demonstrations. 

To evaluate our approach, we compare against several baselines: GPS \cite{zhu-bhat-2021-generate} (a three-stage pipeline), fine-tuned DialoGPT \cite{zhang2020dialogpt} (for contextual coherence), CoARL \cite{hengle2024intentconditioned} (conversational strategy), and the state-of-the-art HiPPrO \cite{padhi-etal-2025-counterspeech} (multi-attribute prefix learning). Furthermore, we evaluate six foundational architectures, encoder-decoders (\textbf{BART-Large} \cite{lewis2020bart}, \textbf{FLAN-T5-XL} \cite{chung2024scaling}) and decoder-only models (\textbf{GPT2-XL}, \textbf{Llama-3.1-Instruct-8B} \cite{dubey2024llama}, \textbf{Mistral-Instruct-7B} \cite{jiang2023mistral7b}, and \textbf{Qwen-2.5-Instruct-7B} \cite{qwen2025qwen25technicalreport}), under zero-shot (ZS) without task specific training, and few-shot (FS) utilizing five in-context exemplars. \textcolor{black}{We also evaluate retrieval-based (RB) standard RAG baselines that conditions generation on the top retrieved exemplar (including its category and evidence), providing single-pass models with equivalent information access to explicitly isolate the performance gains of our multi-agent decomposition (Appendix \ref{prompt_strategies}).} Furthermore, for supervised fine-tuning (SFT) we implement Low-Rank Adaptation (LoRA) \cite{hu2021loralowrankadaptationlarge} fine-tuning with rank 16 on all model architectures.

\vspace{-1mm}
\subsection{Evaluation Metrics}
Evaluating counterspeech generation is challenging due to its context-dependent nature and multiple valid responses for a single HS instance \cite{chung2024understanding}. Our evaluation framework employs multi-dimensional metrics across lexical, semantic, and counterspeech-specific dimensions. We measure lexical similarity using \textbf{METEOR (M)} \cite{banerjee2005meteor}, which incorporates synonym matching, and also report \textbf{ROUGE} scores \cite{lin2004rouge}. Semantic relevance is captured through \textbf{BERTScore (BS)} \cite{zhang2019bertscore} using contextualized embeddings, and \textbf{Cosine Similarity (CoSim)} \cite{reimers2019sentence} between sentence embeddings. We assess linguistic quality through \textbf{Repetition Rate (R)} \cite{cettolo2014repetition} (measuring redundancy via unique token ratios), \textbf{Diversity (Div)} (ratio of unique to total tokens), and \textbf{Novelty (Nov)} (lexical divergence from reference text). \textbf{Toxicity (T)} is measured using Detoxify \cite{hanu2020detoxify} to ensure respectful counterspeech. \textcolor{black}{\textbf{Category Accuracy (CatAcc)} verifies both correct hate speech classification and proper counterspeech strategy execution. We require a strict three-way match: the dataset gold label, the model's predicted hate category, and the corresponding counter-strategy detected in the final response by an independent $RoBERTa_{Large}$ \cite{liu2019robertarobustlyoptimizedbert} classifier.} \textcolor{black}{\textbf{Factuality Score (FSc)} assesses factual correctness using BART-MNLI \cite{Chen_2018} to verify if generated claims are supported by the retrieved gold evidence. Crucially, to fairly measure inherent hallucination, non-retrieval baselines generate responses without access to FIRE's evidence, but all model outputs are evaluated against the same gold facts.}

\section{Experimental Results}
This section presents a comprehensive empirical analysis that evaluates the effectiveness of \texttt{FIRE} in comparison to existing state-of-the-art techniques.

\subsection{Quantitative Results}
\hyperref[tab:results]{Table~\ref{tab:results}} presents the quantitative evaluation across all metrics, where \texttt{FIRE} achieves the best performance across nine of thirteen categories.

\paragraph{Semantic and Lexical Alignment.} \texttt{FIRE} demonstrates superior semantic understanding and contextual relevance while maintaining effective surface-level matching. It leads in all deep semantic metrics, achieving the highest BERTScore (0.886), METEOR (0.247), and CoSIM (0.574), which indicates strong semantic alignment and lexical matching. These gains are supported by competitive ROUGE scores (R1: 0.294, R2: 0.071, RL: 0.197), confirming that the model retains key content while improving semantic depth.

\paragraph{Response Quality and Safety.} \texttt{FIRE} produces content that is significantly safer, more natural, and linguistically diverse than baselines. This is evidenced by a marked improvement in safety, with Toxicity scores dropping to 0.016 (11.1\% lower). The responses are also less repetitive and more varied, indicated by a Repetition Rate of 0.118 (7.1\% lower than the best baseline), alongside increases in Novelty (0.730, 9.1\% higher) and Diversity (0.873, 1.3\% higher), reflecting a richer vocabulary.

\paragraph{Strategic and Factual Integrity.} We observe a substantial boost in Category Accuracy (0.702, 11.1\% higher), reflecting better strategy alignment. Most significantly, the model demonstrates superior reliability with a Factual Score of 0.969 (12.2\% higher), validating its ability to maintain factual accuracy without sacrificing coherence.

\paragraph{Overall Performance.} 
\textcolor{black}{Despite using smaller agents ($<$2B), \texttt{FIRE} proves highly effective against baselines; its sequential activation of two 1.7B agents reduces peak VRAM by 75\% ($\sim$4GB vs. $\sim$16GB for monolithic 8B models) while maintaining comparable end-to-end inference latency (Appendix \ref{app:efficiency}).} Against SFT models, which outperform ZS, FS, and RB approaches, \texttt{FIRE} achieves superior performance across key metrics (see Appendix \ref{stat_test} for significance testing), demonstrating effectiveness in generating factually accurate, and contextually appropriate responses.

\subsection{Ablation Study}
\label{sec:ablation_study}
Our ablation study assesses the effects of each component in \texttt{FIRE} (\hyperref[tab:results]{Table~\ref{tab:results}}). To understand the importance of websearch tool in our framework we remove it, which results in a significant drop in Factual Score (FSc) from 0.969 to 0.832 (14.1\% decrease), demonstrating that availability of supporting information is critical for maintaining factual accuracy in counterspeech. Additionally, Category Accuracy drops from 0.702 to 0.664 (5.4\% decrease). When the memory is removed, the category accuracy falls to 0.599 (14.5\% below full model), while CoSIM drops from 0.574 to 0.490 (14.6\% decrease), indicating reduced contextual coherence. BERTScore decreases from 0.886 to 0.874, and Repetition Rate increases from 0.118 to 0.152, suggesting less diverse and more repetitive outputs. \textcolor{black}{When the HSA is removed, simulating a single-pass scenario with equivalent tool and memory access, FSc and CatAcc drop by 36.0\% and 27.7\%. This proves multi-agent decomposition is functionally necessary to manage cognitive load. Overall, each \texttt{FIRE} component is crucial, web search drives factuality, memory ensures coherence, and the HSA enables targeted strategy. Furthermore, Appendix \ref{app:error_cascading}'s purity score distribution proves \texttt{FIRE} robustly mitigates HSA error cascading by resolving ambiguous classifications before generation.}

\begin{table}[t!]
\begin{center}
\resizebox{.6\columnwidth}{!}{%
\begin{tabular}{l|c|c|c|c} % Removed one 'c' for the last column
\toprule
\multicolumn{1}{l|}{\textbf{Models on comparison}} & \multicolumn{4}{c}{\textbf{Metrics}} \\ % Adjusted for 4 columns
\midrule
 & \textbf{ICS $\uparrow$} & \textbf{Ad $\uparrow$} & \textbf{CoRl $\uparrow$} & \textbf{ArgE $\uparrow$} \\ % Removed CA column
\midrule
\texttt{FIRE} vs Llama-3.1-8b (SFT) & $0.91$ & $0.89$ & $0.86$ & $0.93$ \\
\texttt{FIRE} vs HiPPrO & $0.96$ & $0.94$ & $0.95$ & $0.97$ \\
\texttt{FIRE} vs CoARL & $0.98$ & $0.96$ & $0.96$ & $0.98$ \\
\bottomrule
\end{tabular}%
}
\end{center}
\vspace{-2mm}
\caption{Results of the human evaluation study, where responses generated by \texttt{FIRE} are shown against those produced by (a) Llama-3.1-8b-instruct, (b) HiPPrO, and (c) CoARL. The results are reported in terms of \text{Win Rate \%}, indicating the \% of instances where \texttt{FIRE} outperforms the respective baselines.}
\label{tab:human_eval}
\vspace{-3mm}
\end{table}

\subsection{Human Evaluation}
\label{sec:human_evaluation_section}
Previous studies \cite{jones2024multi, hengle-etal-2025-cseval} highlight that automatic metrics often fail to capture the real-world effectiveness of counterspeech, motivating the need for human evaluation. We therefore conducted a dedicated human study comparing responses generated by \texttt{FIRE} against strong baselines (LLaMA-3.1-8B-Instruct (SFT), HiPPrO, CoARL). We followed \cite{hengle2024intentconditioned} for our human evaluation. Evaluation was performed on a randomly sampled (seed = 2) subset of model outputs, with balanced coverage across hate categories. \textcolor{black}{A total of $30$ expert annotators (aged 20–35, balanced gender) with relevant backgrounds ranked paired responses from 1 (Best) to 4 (Worst) across four criteria: Independent Counterspeech (ICS), Adequacy (Ad), Contextual Relevance (CoRl), and Argumentative Effectiveness (ArgE) (Appendix \ref{app:human_eval_rubric}).} Results are reported using win rates, representing the fraction of instances where \texttt{FIRE} was preferred over the baseline. As shown in \hyperref[tab:human_eval]{Table \ref{tab:human_eval}} , \texttt{FIRE} consistently outperforms all comparison models across every metric. Against best baseline method LLaMA-3.1-8B-Instruct SFT, \texttt{FIRE} achieves strong win rates across ICS (0.91), Ad (0.89), CoRl (0.86), and ArgE (0.93), indicating clear human preference for its fluency, contextual grounding, and argumentative strength.

\begin{table}[!t]
\small
\centering
\begin{tabular}{p{0.15\columnwidth} p{0.7\columnwidth}}
\toprule
\textbf{Type} & \textbf{Content} \\
\midrule
\textbf{Hate Speech} & No muslims are compatible with any Western society. \\
\midrule
\textbf{Model} & \textbf{Generated Counterspeech} \\
\midrule
\textbf{Llama-3.1-8b-instruct} & Muslims are a diverse group and \textcolor{red}{integration varies across countries}. Many Muslims contribute positively to Western societies. \\
\midrule
\textbf{\texttt{FIRE} (Ours)} & A Muslim leader like \textcolor{blue}{London Mayor Sadiq Khan} has integrated into Western societies while upholding Islamic values, demonstrating that \textcolor{blue}{compatibility is possible through mutual respect and dialogue}. Diversity enriches societies, and stereotypes ignore the complexity of individual experiences. Understanding Islamophobia requires recognizing that \textcolor{blue}{cultural and religious differences foster mutual growth}, not division. \\
\bottomrule
\end{tabular}

\caption{Error analysis between the response generated by our method \texttt{FIRE} vs the best baseline.}
\label{tab:error_analysis}
\end{table}

\section{Error Analysis}
% \hyperref[tab:error_analysis]{Table \ref{tab:error_analysis}} shows the qualitative disparity in counterspeech between best performing baseline Llama-3.1-8B-Instruct and \texttt{FIRE}. While responding to the claim `No Muslims are compatible with any Western society', the baseline offers a hedged rebuttal; its statement that `integration varies across countries' unintentionally validates the stereotype by implying conditional legitimacy. In contrast, \texttt{FIRE} employs a strategically grounded structure that directly invalidates the hate claim through verifiable counter-evidence. By citing a specific example `Muslim leader like London Mayor Sadiq Khan' and highlighting how such figures have `integrated into Western societies while upholding Islamic values,' the model replaces abstract defensiveness with concrete proof. Furthermore, \texttt{FIRE} actively reframes the concept of `compatibility' not as a clash, but as a dynamic achieved through `mutual respect and dialogue,' thereby demonstrating a superior capacity for constructive recontextualization.

\hyperref[tab:error_analysis]{Table \ref{tab:error_analysis}} shows the qualitative disparity in counterspeech between best performing baseline Llama-3.1-8B-Instruct and \texttt{FIRE}. Responding to the claim `No Muslims are compatible with any Western society', the baseline offers a hedged rebuttal; its statement that `integration varies across countries' unintentionally validates the stereotype by implying conditional legitimacy. Conversely, \texttt{FIRE} directly invalidates the universal generalization through concrete counter-evidence (`London Mayor Sadiq Khan'). By reframing compatibility as `mutual respect', \texttt{FIRE} moves beyond the baseline's passive neutrality to provide a constructive, factually grounded counterspeech.

\section{Conclusion}
In this work, we introduce \texttt{FIRE}, a hierarchical agentic framework, and \texttt{FactualCS}, a dataset targeting different hate speech categories. By decomposing generation into intent analysis and grounded response synthesis, \texttt{FIRE} enables categorically precise and factually supported interventions. Using compact ($<$2B) models without any finetuning, \texttt{FIRE} substantially outperforms established baselines and matches larger state-of-the-art LLMs, demonstrating that collaborative, reasoning-centric agents provide an efficient and effective approach to safe, context-aware counterspeech.

% In this work, we introduced \texttt{FIRE}, a hierarchical agentic framework, alongside \texttt{FactualCS}, a novel dataset designed to address the nuances of varied hate speech categories. By decomposing the generation task into a two-stage process, intent analysis followed by grounded response synthesis, our approach ensures that interventions are both categorially precise and factually supported. Despite utilizing compact models ($<$2B parameters) without fine-tuning, \texttt{FIRE} significantly outperforms established baselines and matches larger state-of-the-art LLMs. These results demonstrate that collaborative, reasoning focused agents offer a highly efficient and effective pathway for generating safe, context-aware counterspeech.

\section*{Limitation}

Our research presents several limitations that warrant consideration. Firstly, while \texttt{FactualCS} targets five distinct categories of abuse, the dataset is not exhaustive, potentially limiting generalization to other forms of hate speech or intersectional attacks not covered within our $14$ target communities. Secondly, our reliance on small language models ($<$2B), though efficient, may constrain the depth of reasoning compared to larger state-of-the-art systems, particularly when decoding highly ambiguous linguistic nuances. Additionally, the agentic workflow introduces a dependency on external tools; web search failures can propagate errors into the final response, compromising factual grounding. Further, We also acknowledge that the annotation team was relatively small and consisted primarily of researchers with technical expertise; although this expertise was valuable for applying the operational taxonomy and evidence-retrieval criteria, the annotators did not necessarily represent the lived experiences or cultural backgrounds of all target communities covered by FactualCS. Consequently, culturally specific, implicit, coded, or community-dependent expressions of hate may remain underrepresented or may be subject to interpretation bias. \textcolor{black}{Furthermore, our current evaluation is restricted to English-language datasets, which may limit applicability to the diverse linguistic and cultural nuances of hate speech in other regions.} Lastly, while we demonstrate high factual accuracy, our evaluation primarily focuses on immediate response quality rather than long-term conversational impact or the potential for conflict escalation. \textcolor{black}{Future work could address these gaps by expanding the taxonomy of hate types, adapting our inherently language-agnostic architecture to diverse cultural contexts via multilingual models and localized search APIs, and incorporating longitudinal studies on user interactions.}
\section*{Ethics Statement}
We recognize the sensitivity required in addressing online hate speech and acknowledge the ethical and moral complexities inherent in conducting research in this area. This initiative serves as an initial attempt to compile a comprehensive and varied collection of counterspeech responses for each instance of hate speech encountered. We understand that algorithms developed for automated counterspeech may generate responses that fail to accurately convey the intended meanings, highlighting the urgent need to better integrate real-world knowledge into these systems. Despite the potential of generative algorithms, there remains a critical necessity for a robust and diverse database of counterspeech to ensure consistently favorable outcomes. Furthermore, while fully operational counterspeech algorithms have yet to be realized, organizations such as United Against Hate play a crucial role in mitigating the prevalence of hate speech in online environments.

% \section{Acknowledgments}
% We extend our gratitude to the central HPC facility (Padum) at IIT Delhi for computing. We also sincerely thank Logically and Anusandhan National Research Foundation (DST-NSF Grant: DST/INT/USA/NSF-DST/Tanmoy/P-2/2024) for financial support. Tanmoy acknowledges the support of Rajiv Khemani Young Faculty Chair Professorship in Artificial Intelligence. 

\bibliographystyle{IEEEtranN}
\small
\bibliography{anthology}
\newpage
\appendix
\section{Appendix}
\label{sec:appendix-dataset}
\subsection{Hatespeech Selection}
\label{sec:hatespeech_selection}
To construct the \texttt{FactualCS} dataset, we aggregated hate speech text from four established datasets to capture diverse dynamics of online abuse: \textit{IntentCONANv2} \cite{hengle2024intentconditioned} (conversational abuse), \textit{HatEval} \cite{hateval} (identity-targeted attacks), \textit{ETHOS} \cite{mollas2022ethos} (discrimination and violence), and the \textit{Gab Hate Corpus} \cite{kennedy2018gab} (extreme dehumanization). Our selection process involved multiple filtering stages to ensure dataset quality. First, we removed redundant hate speech instances across the five datasets, reducing the initial collection to $8,000$ unique samples. Next, we filtered out samples constrained by brevity (those under 5 words) that lacked sufficient context for meaningful analysis, retaining only self-expressive hate speech that clearly articulated harmful intent or targeted content. We extracted 2,384, 1,500, 550, 350, samples from IntentCONANv2, HatEval, ETHOS, Gab Hate Corpus respectively.

\subsection{Annotator Demographics}
Six annotators volunteered in the annotation process, comprising 4 male and 2 female annotators. All annotators possess strong computational backgrounds with expertise in natural language processing and hate speech analysis. The team includes researchers with prior publications in areas of content moderation, computational social science, and counter-narrative generation, ensuring they brought both technical proficiency and domain knowledge to the annotation task.

\subsection{Annotation Process}
\label{sec:data_annotation}
\subsubsection{Pilot Annotation Phase}
Before undertaking full-scale annotation, we conducted a pilot phase with 500 hate speech samples to establish annotation protocols and achieve inter-annotator consensus. During the proprietary phase, all six annotators first independently annotated each sample, after which they collaboratively reviewed and refined the annotations through mutual discussion and deliberation. For each hate speech instance, annotators identified and documented: (1) the hate speech text, (2) hate speech type, (3) reasoning behind choosing this hate type, (4) target group, (5) contextual query, (6) supporting evidence (when applicable), (7) counterspeech response, (8) source URL, (9) source date, and (10) source author. We classified hate speech into five categories: misinformation, stereotype, conspiracy theory, dehumanizing language, and non-factual claims. For hate speech types requiring factual verification specifically misinformation, stereotypes, and conspiracy theories, annotators conducted web searches to locate credible evidence and documented all relevant source details through manual annotation. Initial inter-annotator agreement (IAA) scores revealed substantial disagreement among annotators. The primary sources of divergence included: (1) distinguishing between stereotype and misinformation, (2) determining the threshold for dehumanizing language versus harsh criticism, and (3) identifying when stereotypes required explicit factual evidence versus contextual refutation. Through iterative discussion sessions and collaborative review of contentious cases, annotators developed a shared understanding of category boundaries and annotation standards. These deliberations led to the formulation of comprehensive annotation guidelines (detailed in Appendix \ref{procedure_annotatioin}), which operationalized each hate speech type with clear definitional criteria and decision rules. By the conclusion of the pilot phase, IAA scores exceeded 0.9 across all annotation dimensions (See Table \ref{tab:gold_annotation}), and the 500 annotated samples formed our gold-standard reference dataset.

\subsubsection{Main Annotation Phase}
Following the pilot phase, the six annotators proceeded to annotate the remaining dataset using the established guidelines detailed in Appendix \ref{procedure_annotatioin}. All annotation components: hate speech type identification, reasoning behind hate speech type choice, target group detection, query formulation, and evidence collection, were completed manually to ensure high-quality labels. Inter-annotator agreement during the main phase remained consistently high around 91.55\%, with scores reported in Table \ref{tab:interm_annotation}, demonstrating that the guidelines successfully transferred to the broader dataset. However, generating counterspeech responses for thousands of samples proved prohibitively time-intensive for manual annotation. To address this efficiency challenge while maintaining quality, we employed Gemini-2.0-Flash \cite{team2023gemini} to generate counterspeech responses. We provided the model with few-shot examples from our $500$ sample gold-standard dataset along with the manually annotated hate speech attributes (type, reasoning, group, query, evidence) as context for generation. This hybrid approach allowed us to scale counterspeech generation while grounding outputs in carefully curated examples. To validate counterspeech quality, annotators evaluated a random sample of generated responses across multiple dimensions including appropriateness, factual accuracy, and persuasiveness. Inter-annotator agreement scores for counterspeech quality assessment are presented in Table \ref{tab:counter_annotation}, indicating that the generated counterspeech met our quality standards and exhibited consistency with the manually authored examples from the pilot phase.

\begin{table*}[t!]
\centering
{
\setlength{\tabcolsep}{1mm}
\begin{tabular}{l|c|c|c|c|c|c}
\toprule
 & \textbf{Annotator 1} & \textbf{Annotator 2} & \textbf{Annotator 3} & \textbf{Annotator 4} & \textbf{Annotator 5} & \textbf{Annotator 6} \\
\midrule
Annotator 1 & 1.0000 & 0.9220 & 0.9500 & 0.9260 & 0.9420 & 0.9540 \\
Annotator 2 & 0.9220 & 1.0000 & 0.9440 & 0.9240 & 0.9240 & 0.9520 \\
Annotator 3 & 0.9500 & 0.9440 & 1.0000 & 0.9480 & 0.9520 & 0.9760 \\
Annotator 4 & 0.9260 & 0.9240 & 0.9480 & 1.0000 & 0.9280 & 0.9560 \\
Annotator 5 & 0.9420 & 0.9240 & 0.9520 & 0.9280 & 1.0000 & 0.9560 \\
Annotator 6 & 0.9540 & 0.9520 & 0.9760 & 0.9560 & 0.9560 & 1.0000 \\
\bottomrule
\end{tabular}
}
\caption{Inter-annotator Agreement Coefficients for 500 samples in Pilot Annotation Phase}
\label{tab:gold_annotation}
\end{table*}

\begin{table*}[t!]
\centering
\scalebox{1}
{
\setlength{\tabcolsep}{1mm}
\begin{tabular}{l|c|c|c|c|c|c}
\toprule
 & \textbf{Annotator 1} & \textbf{Annotator 2} & \textbf{Annotator 3} & \textbf{Annotator 4} & \textbf{Annotator 5} & \textbf{Annotator 6} \\
\midrule
Annotator 1 & 1.0000 & 0.8860 & 0.9200 & 0.9000 & 0.9120 & 0.9300 \\
Annotator 2 & 0.8860 & 1.0000 & 0.9140 & 0.8900 & 0.8900 & 0.9240 \\
Annotator 3 & 0.9200 & 0.9140 & 1.0000 & 0.9240 & 0.9280 & 0.9540 \\
Annotator 4 & 0.9000 & 0.8900 & 0.9240 & 1.0000 & 0.9000 & 0.9300 \\
Annotator 5 & 0.9120 & 0.8900 & 0.9280 & 0.9000 & 1.0000 & 0.9300 \\
Annotator 6 & 0.9300 & 0.9240 & 0.9540 & 0.9300 & 0.9300 & 1.0000 \\
\bottomrule
\end{tabular}
}
\caption{Inter-annotator Agreement Coefficients in Main Annotation Phase for Hate Speech Type, Reasoning, Target Group, Query, and collected evidence}
\label{tab:interm_annotation}
\end{table*}

\begin{table*}[t!]
\centering
{
\setlength{\tabcolsep}{1mm}
\begin{tabular}{l|c|c|c|c|c|c}
\toprule
 & \textbf{Annotator 1} & \textbf{Annotator 2} & \textbf{Annotator 3} & \textbf{Annotator 4} & \textbf{Annotator 5} & \textbf{Annotator 6} \\
\midrule
Annotator 1 & 1.0000 & 0.8960 & 0.9180 & 0.9080 & 0.9140 & 0.9240 \\
Annotator 2 & 0.8960 & 1.0000 & 0.9180 & 0.9000 & 0.8980 & 0.9200 \\
Annotator 3 & 0.9180 & 0.9180 & 1.0000 & 0.9180 & 0.9240 & 0.9380 \\
Annotator 4 & 0.9080 & 0.9000 & 0.9180 & 1.0000 & 0.9060 & 0.9240 \\
Annotator 5 & 0.9140 & 0.8980 & 0.9240 & 0.9060 & 1.0000 & 0.9220 \\
Annotator 6 & 0.9240 & 0.9200 & 0.9380 & 0.9240 & 0.9220 & 1.0000 \\
\bottomrule
\end{tabular}
}
\caption{Inter-annotator Agreement Coefficients for generated counterspeech}
\label{tab:counter_annotation}
\end{table*}

\begin{table*}[t]
\centering
\small
\setlength{\tabcolsep}{3pt}

% \begin{tabular}{lllll}
\resizebox{\textwidth}{!}{
\begin{tabular}{p{2.2cm} p{5.5cm} p{4.5cm} p{5cm} p{1.2cm}}

% {p{2.4cm} p{3.5cm} p{2.8cm} p{3.8cm} p{1.2cm}}
\toprule
\textbf{HS Type} &
\textbf{Description} &
\textbf{Example} &
\textbf{Ideal Response Type} &
\textbf{Factual?} \\
\midrule

Misinformation &
Relies on false or misleading factual claims about a group &
Immigrants increase crime rates. &
Fact-based rebuttal using data, statistics, authoritative sources &
Yes \\
\midrule

Conspiracy &
Misinformation mixed with large-scale conspiracy framing &
Jews control the world’s banks. &
Debunk narrative patterns, use cognitive inoculation strategies &
Yes \\
\midrule

Stereotype &
Cultural or historical generalizations applied to a group &
Women can’t handle leadership. &
Reframing, factual correction, highlighting counterexamples &
Yes \\
\midrule

Dehumanizing &
Use of animalistic or violent metaphors &
Refugees are parasites. &
Moral, legal, and social norm reminders &
No \\

\midrule
Non-factual &
Hatred or disgust without factual claims &
Muslims are disgusting. &
Empathy-based counterspeech, reinforcing social norms &
No \\
\bottomrule
\end{tabular}}
\caption{Hate Speech Types based on Factuality, Response Strategies, and Factual Rebuttal Requirements}
\label{tab:hs_taxonomy}
\end{table*}

\subsection{Procedure and Annotation Criteria}
\label{procedure_annotatioin}
Before beginning the annotation process, all annotators thoroughly reviewed comprehensive resources on addressing online harassment and counterspeech strategies, including the field guide on ``addressing online harassment'' \footnote[5]{https://onlineharassmentfieldmanual.pen.org/}. This preparatory phase involved extensive discussions with the annotators to deepen their understanding of hate speech taxonomy and effective counter-narrative generation. These dialogues ensured that the annotators were well-equipped with the necessary knowledge and context, enabling them to effectively contribute to the project with a shared conceptual framework.

\subsubsection{Hate Speech Type Definitions and Criteria}
The annotation process required annotators to classify each hate speech instance into one of five mutually exclusive categories. The following criteria were adhered to consistently throughout the annotation process.

\paragraph{Misinformation:}
Misinformation refers to hate speech containing factually incorrect or misleading claims that can be empirically verified against credible sources. This category encompasses false statements presented as facts that target or disparage specific groups, regardless of whether the misinformation was spread intentionally or unintentionally. The annotation criteria for this:
\begin{itemize}
    \item The hate speech must contain at least one specific, verifiable factual claim that is demonstrably false.
    \item The false claim must directly contribute to the hateful characterization of the target group.
    \item Annotators must identify the precise false claim within the hate speech.
    \item Evidence contradicting the misinformation must be drawn from authoritative sources such as peer-reviewed research, government statistics, reputable news organizations, or fact-checking institutions.
    \item The evidence must be current and contextually relevant to the claim being refuted.
\end{itemize}

\paragraph{Stereotype:}
Stereotype refers to hate speech that perpetuates oversimplified, generalized, or prejudiced characterizations of social groups. These are fixed, commonly held beliefs about groups that reduce individual complexity to monolithic attributes, often negative or limiting in nature. The annotation criteria for this:

\begin{itemize}
    \item The hate speech must attribute characteristics, behaviors, or traits to an entire group based solely on group membership.
    \item The characterization must be reductive, denying individual variation within the group.
    \item Evidence to counter stereotypes should include: statistical data showing group diversity, research findings on within-group variation, documented counter-examples, or scholarly work debunking the stereotype.
    \item For cultural or behavioral stereotypes, evidence may include anthropological studies, sociological research, or representative examples demonstrating group heterogeneity.
\end{itemize}

\textbf{How Stereotype is different from Misinformation?} Stereotypes make generalized claims about group characteristics; misinformation makes specific false factual claims. Stereotypes may not be empirically testable in the same way (e.g., ``X group is lazy'' vs. ``X group has 90\% unemployment''). When a stereotype includes a verifiable false statistic, classify as misinformation.

\paragraph{Conspiracy Theory:} Conspiracy theory refers to hate speech promoting unfounded explanatory narratives that attribute events, policies, or social phenomena to secret plots by powerful actors, typically targeting specific groups as the alleged conspirators. These narratives reject mainstream explanations in favor of hidden malevolent coordination. The annotation criteria for this:

\begin{itemize}
  \item The hate speech must allege a secret, coordinated plan by a specific group to achieve nefarious goals.
  \item The conspiracy must lack credible evidentiary support and reject established explanations.
  \item The alleged conspiracy must involve extraordinary claims of coordination, secrecy, or power.
  \item Evidence must demonstrate: (a) the lack of credible support for the conspiratorial claim, and (b) when possible, factual explanations for the events or phenomena in question.
  \item Evidence sources should include: investigative journalism, expert analysis, academic debunking, or fact-checking reports.
\end{itemize}

\paragraph{Dehumanizing Language:} Dehumanizing language refers to hate speech that depicts individuals or groups as subhuman, animalistic, objects, or fundamentally less than human. This category focuses on linguistic markers that strip target groups of human dignity, moral status, or basic humanity. The annotation criteria for this:

\begin{itemize}
  \item The hate speech must use language that explicitly or implicitly denies the full humanity of the target group.
  \item Dehumanization can take multiple forms: animalistic (e.g., vermin, animals), mechanistic (e.g., tools, objects), biological (e.g., disease, infestation), or exclusionary (e.g., not deserving of human rights).
  \item This category focuses on the nature of the language used rather than factual claims.
  \item Evidence is typically not required for this category, as it concerns linguistic framing rather than empirical claims.
  \item However, contextual information about why such language is harmful and its historical usage may be included.
\end{itemize}

\paragraph{Non-Factual Claims:} 
Non-factual claims refer to hate speech expressing opinions, value judgments, moral condemnations, or abstract assertions that cannot be empirically verified through evidence but still promote harmful attitudes, prejudice, or discrimination toward target groups. The annotation criteria for this:

\begin{itemize}
  \item The hate speech must express subjective evaluations, opinions, or value judgments rather than factual claims.
  \item The statements are not amenable to fact-checking but nonetheless promote hatred or prejudice. These may include: aesthetic judgments, moral condemnations, predictions about the future, personal preferences framed as group characteristics, or abstract value-laden assertions.
  \item Counterspeech for this category typically involves explaining why such rhetoric is harmful, its potential effects, or providing counter-narratives that promote inclusive values.
\end{itemize}

\subsubsection{Additional Annotation Requirements}
\paragraph{Reasoning Behind Hate Speech Type Choice}
For each classification decision, annotators must provide explicit reasoning explaining why the hate speech was assigned to a particular category. This reasoning should reference specific textual elements and explain how they meet the definitional criteria for the chosen category. This metadata helps in decision-making processes and facilitates review and quality control.

\paragraph{Target Group Identification}
Annotators must identify the specific social group(s) targeted by the hate speech. Target groups may be defined by: race, ethnicity, religion, nationality, gender, sexual orientation, disability status, or other protected characteristics. When multiple groups are targeted, all should be documented.

\paragraph{Query Formulation}
For hate speech types requiring evidence (misinformation, stereotype, conspiracy theory), annotators must formulate clear, focused search queries that would retrieve relevant counter-evidence. Queries should be specific enough to yield targeted results but broad enough to capture relevant sources. Example: For "Muslims commit most terrorist attacks," a query might be "global terrorism statistics by perpetrator group."

\paragraph{Evidence Documentation}
When evidence is required, annotators must document:
\begin{itemize}
    \item Source URL: Complete web address of the evidence source.
    \item Source Date: Publication or last updated date.
    \item Source Author/Organization: Individual author or publishing organization.
    \item Relevant Excerpt: Brief quote or summary of the evidence that directly counters the hate speech claim.
    \item Credibility Assessment: Brief note on why the source is authoritative (e.g., peer-reviewed, government statistics, reputable news organization)
\end{itemize}

\subsubsection{Edge Cases and Decision Rules}
\paragraph{Multiple Category Applicability:}
When hate speech could fit multiple categories, annotators should prioritize based on the most prominent or harmful element. The decision hierarchy typically follows: (1) Dehumanizing language (if explicit dehumanization is present), (2) Conspiracy theory (if elaborate coordination is alleged), (3) Misinformation (if specific false facts are central), (4) Stereotype (if generalization is primary), (5) Non-factual claims (as default for subjective assertions).

\paragraph{Ambiguous Cases:}
When categorization is uncertain, annotators should document the ambiguity in the reasoning field and, during the pilot phase, flag for group discussion. In the main phase, annotators should apply the guidelines as consistently as possible and note any uncertainties.

\paragraph{Intersectional Targeting:}
When hate speech targets multiple groups simultaneously (e.g., "Muslim women"), annotators should document all target groups and consider how intersecting identities may compound the harm.

\subsection{\textcolor{black}{\texttt{FactualCS} Taxonomy Distinctions}}
\label{ap:taxonomy_distinctions}
\textcolor{black}{
To ensure robust categorization and address the diverse semantic nature of online abuse, our taxonomy is grounded in established social science literature. Rather than treating hate speech as a homogeneous entity, we define five mutually exclusive categories that act as functionally distinct logical triggers, each necessitating a fundamentally different counterspeech strategy as summarized in Table \ref{tab:hs_taxonomy}. The core distinction relies on separating claims that are empirically verifiable from those based purely on subjective affect or linguistic framing.
}

\textbf{\textcolor{black}{Misinformation versus Stereotype:}} \textcolor{black}{While both categories fall under factual hate, they operate on different levels of specificity. Following the distinctions outlined by \cite{wardle2024conceptual} and \cite{vargas2023socially}, misinformation relies on specific, falsifiable events or fabricated statistics (e.g., ``Immigrants increase crime rates''), requiring a direct, empirical fact-check using authoritative data. In contrast, stereotypes make broad, reductive cultural or historical generalizations about a group's traits or abilities (e.g., ``Women can't handle leadership''). Stereotypes are countered not merely with isolated statistics, but by highlighting group heterogeneity and providing specific counterexamples to reframe the narrative.
}

\textbf{\textcolor{black}{Stereotype versus Non-Factual Hate:}} \textcolor{black}{A critical boundary exists between stereotypes and non-factual hate, addressing potential ambiguities in classification. The distinction relies entirely on the presence of a propositional claim. Stereotypes, despite being generalizations, contain a premise that can be logically debated and factually refuted. Conversely, non-factual hate consists of subjective expressions of disgust, slurs, or opinions without any verifiable assertion (e.g., ``Muslims are disgusting''). Because non-factual hate contains no empirical claim to disprove, attempting to ``fact-check'' it results in a logical mismatch. Therefore, non-factual expressions trigger an entirely different strategic response based on empathy and social norm reinforcement.
}

\textbf{\textcolor{black}{Conspiracy Theories versus Misinformation:}} \textcolor{black}{Incorporating the definitions from \cite{jolley2020exposure}, conspiracy theories are differentiated from standard misinformation by their structural narrative. They allege secret, large-scale coordination by powerful actors (e.g., ``Jews control the world's banks''). Because these narratives often reject mainstream evidence as part of the cover-up, effective counterspeech requires debunking the narrative patterns and utilizing cognitive inoculation strategies, rather than simply presenting contradictory facts.
}

\textbf{\textcolor{black}{Dehumanization versus General Non-Factual Hate:}} \textcolor{black}{Drawing on the framework of \cite{mendelsohn2020framework}, dehumanization is separated from general non-factual hate due to its specific linguistic severity. It utilizes animalistic, mechanistic, or violent metaphors (e.g., ``Refugees are parasites'') to actively strip target groups of their human dignity and moral status. This extreme form of abuse requires counterspeech that explicitly reminds the user of moral, legal, and social norms to re-establish the target's humanity. By strictly delineating these boundaries, \texttt{FactualCS} ensures that models learn the underlying intent of the abuse, allowing for precise, strategy-aligned generation.
}

\textcolor{black}{
Beyond defining these boundaries, it is crucial to justify the specific granularity of our five-category taxonomy compared to coarser or finer alternatives. A coarser categorization scheme (such as a simple binary Hate versus Not-Hate classification) is fundamentally insufficient because it treats distinct logical flaws homogeneously. For example, grouping misinformation with dehumanization forces a model into "logical mismatching," where it might attempt to empirically fact-check a subjective slur or provide a normative moral opinion against a falsifiable statistic. The empirical necessity of this specific decomposition is proven in our ablation study (Section \ref{sec:ablation_study}); when the Hate Speech Analyst is removed ("Without HSA") and the model must generate counterspeech without categorical gatekeeping, the Factual Score severely drops from 0.969 to 0.620 (a 35\% decrease). Conversely, a finer-grained taxonomy would introduce redundant complexity and data sparsity without fundamentally altering the required tool usage or reasoning paths. Our five categories represent the optimal resolution, sufficiently detailed to trigger the correct core reasoning strategy (empirical fact-checking, logical rebuttal, or normative reframing) while remaining computationally efficient for compact agents.
}

\subsection{Advantage of \texttt{FactualCS}}
\label{adav_factualcs}
\texttt{FactualCS} advances counterspeech dataset construction by addressing critical gaps in existing resources and introducing evidence and strategy aware counter-narratives for combating online hate. Unlike prior datasets that primarily emphasize rhetorical or emotional responses, \texttt{FactualCS} grounds counterspeech in verifiable evidence with explicit source attribution and claim–rebuttal linkage, making it effective against misinformation, stereotype, and conspiracy driven hate. Importantly, it is not limited to factual refutation: through a fine-grained hate speech taxonomy, \texttt{FactualCS} enables strategy-specific counterspeech for non-factual and dehumanizing cases, where value-based reframing and rehumanization are more appropriate. Each instance is further enriched with contextual metadata such as target groups, reasoning, and evidence provenance. By aggregating hate speech from multiple established corpora and employing a rigorous hybrid human–AI annotation pipeline with high inter-annotator agreement (>$0.8$), \texttt{FactualCS} achieves both diversity and quality at scale, resulting in a reliable and practically useful resource for hate speech type aware counterspeech generation.

\subsection{Dataset Statistics}
\label{data_stat}
\begin{figure*}[!t]
  \centering
  \small
  \setlength{\belowcaptionskip}{-15pt}
  
  % --- First Row: Dataset Splits ---
  \begin{subfigure}{.4\textwidth} 
    \centering
    \includegraphics[width=0.9\linewidth]{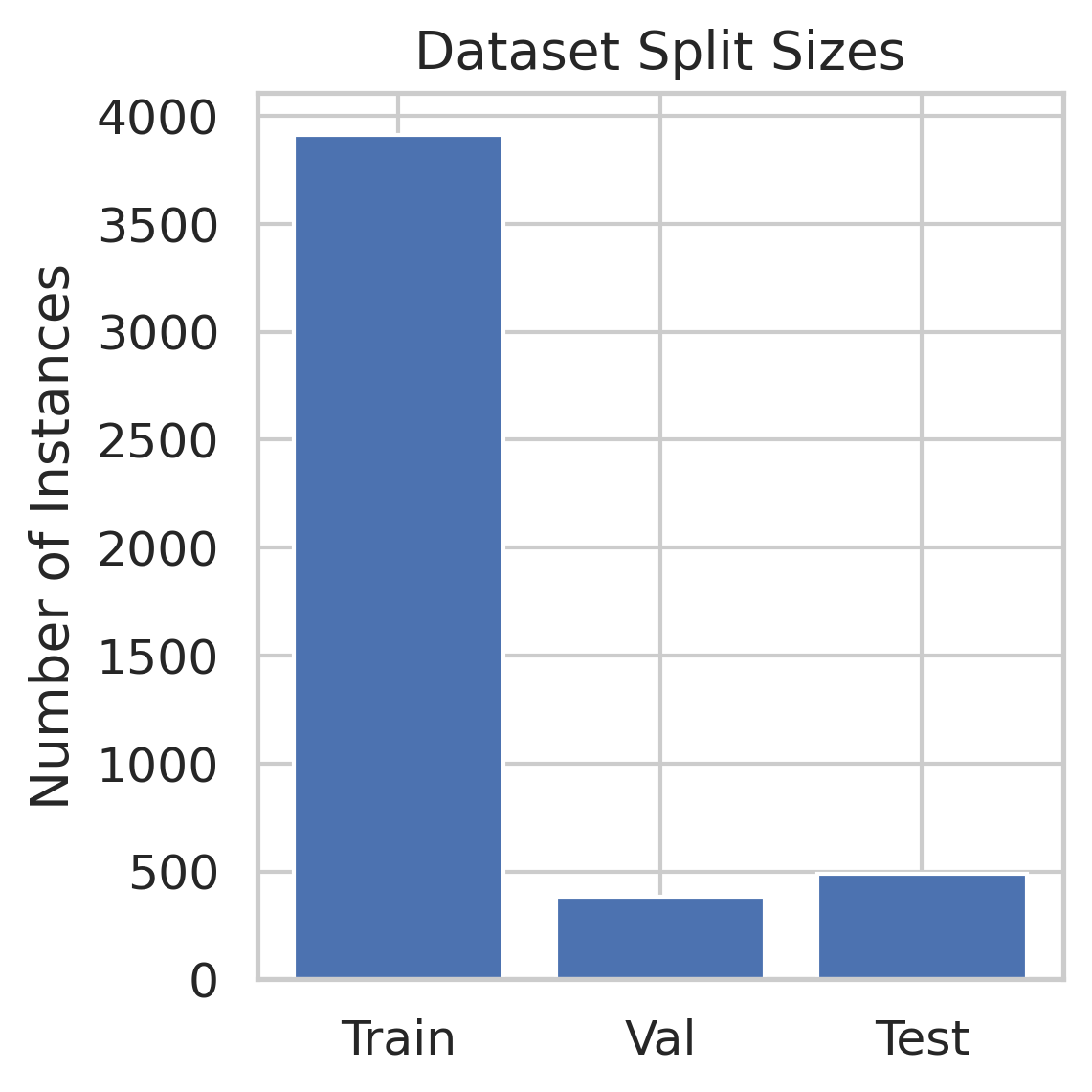}
    \caption{Dataset split sizes across Train, Val, and Test.}
    \label{fig:split_sizes}
  \end{subfigure}\hfill%
  \begin{subfigure}{.48\textwidth} 
    \centering
    \includegraphics[width=0.9\linewidth]{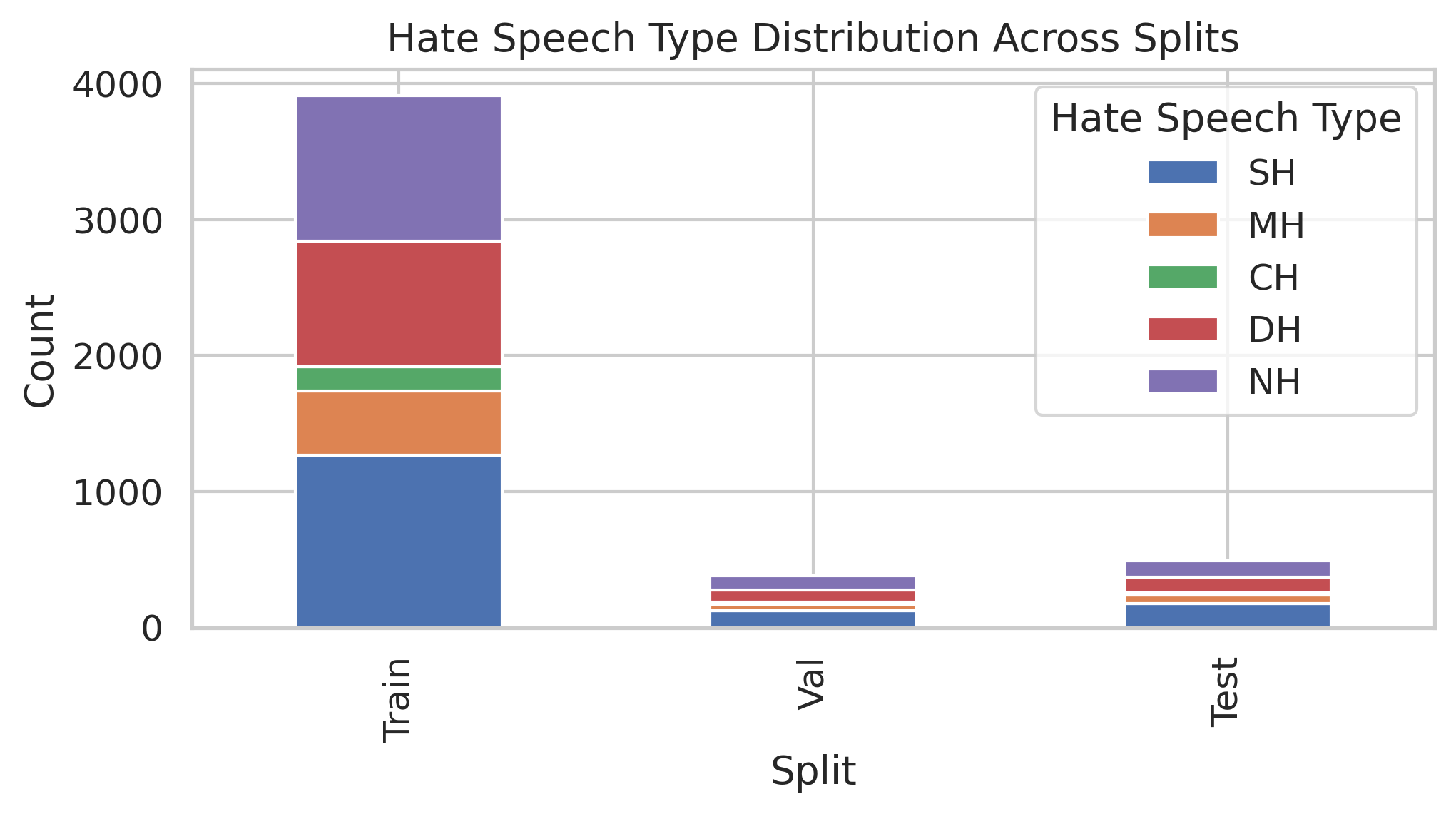}
    \caption{Distribution of Hate Speech types across splits.}
    \label{fig:hs_type_split_stacked}
  \end{subfigure}
  
  \medskip
  \medskip
  
  % --- Middle Row: Detailed Retrieval Analysis (Wide) ---
  \begin{subfigure}{.8\textwidth} 
    \centering
    \includegraphics[width=0.75\linewidth]{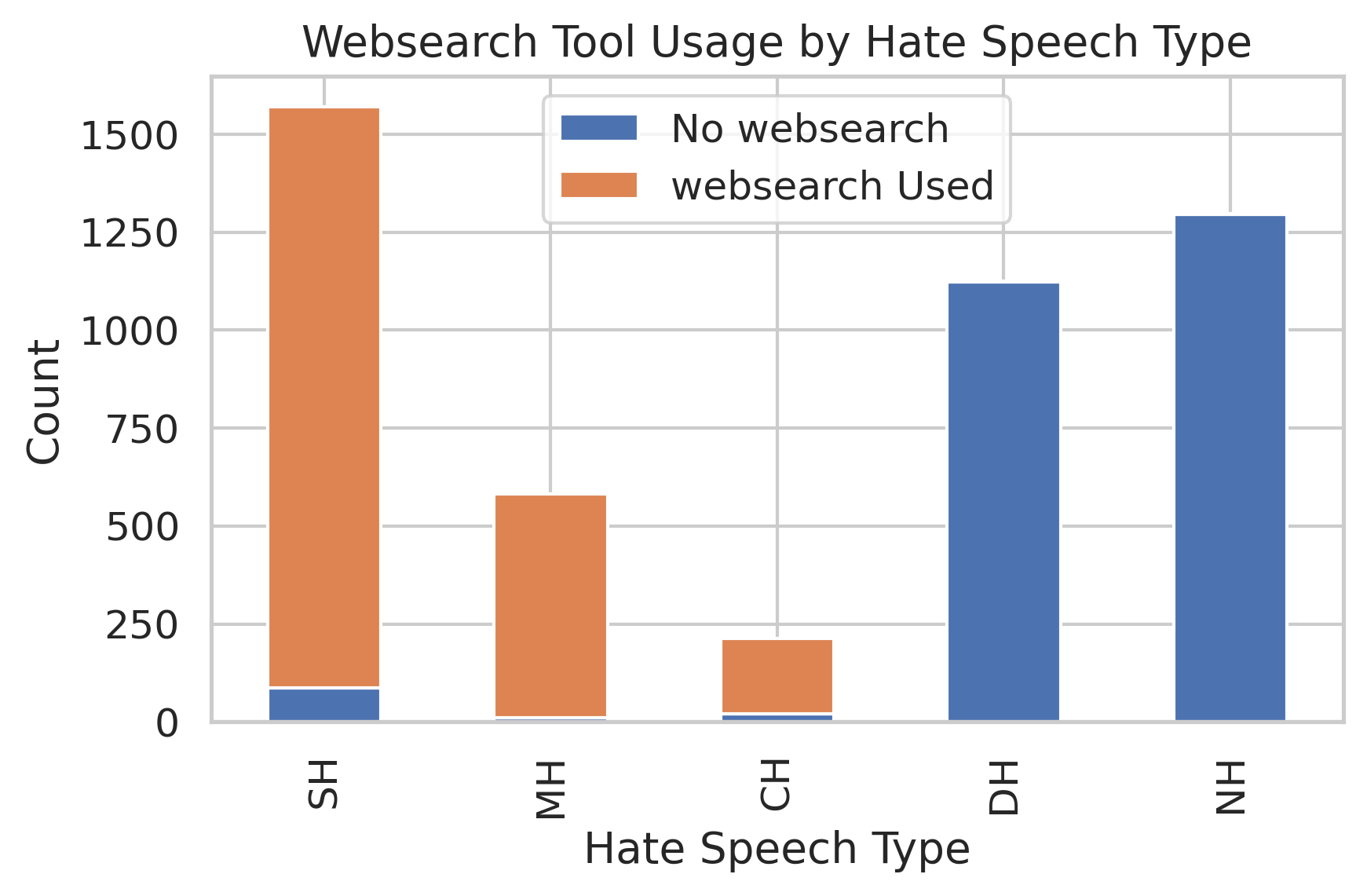}
    \caption{Analysis of Websearch Tool usage conditioned on specific Hate Speech categories.}
    \label{fig:hs_type_vs_retrieval}
  \end{subfigure}
  
  \medskip
  \medskip
  
  % --- Bottom Row: Proportions (Pie Charts) ---
  \begin{subfigure}{.48\textwidth} 
    \centering
    \includegraphics[width=0.8\linewidth]{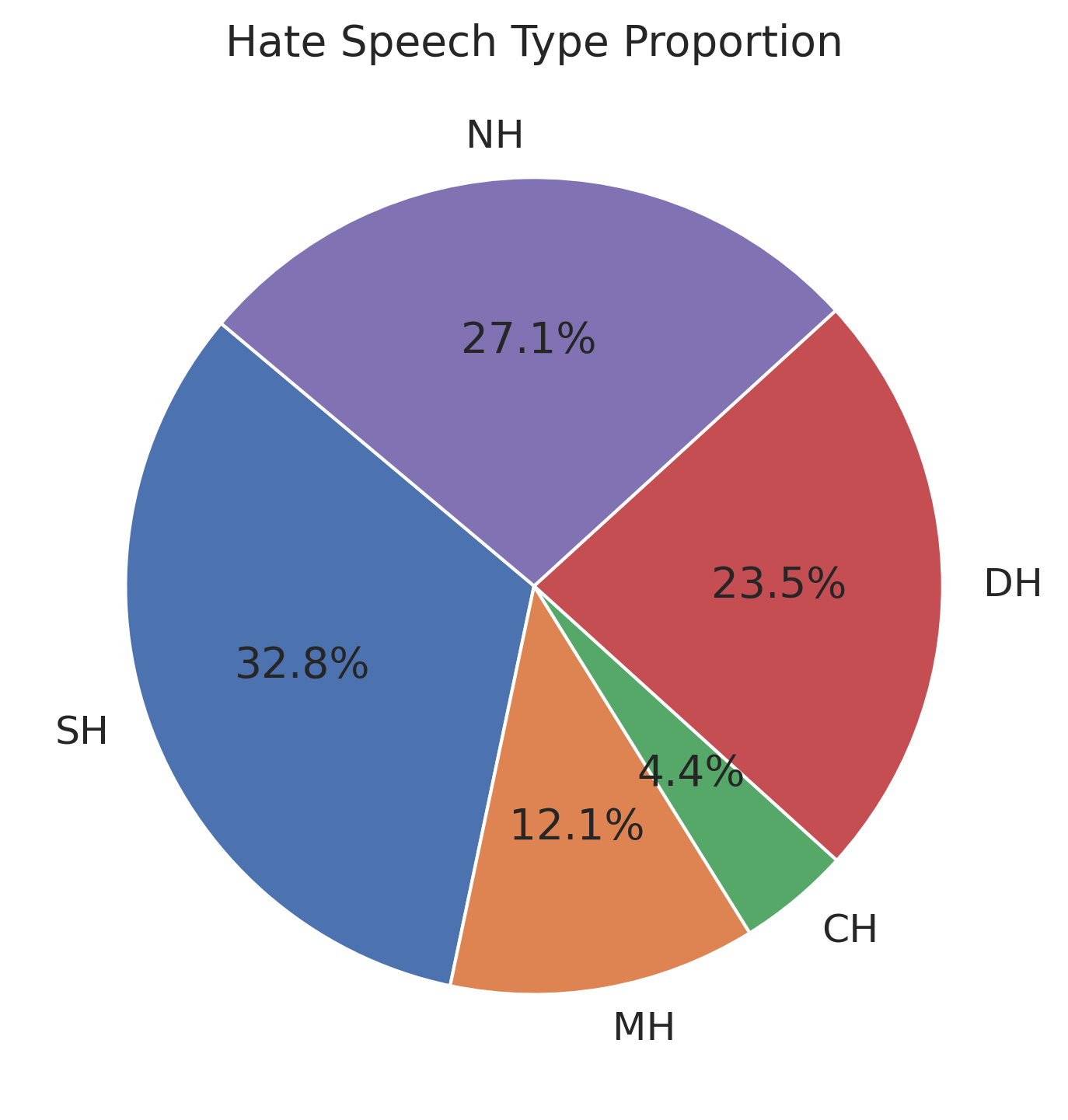}
    \caption{Overall proportion of Hate Speech types.}
    \label{fig:hs_type_pie}
  \end{subfigure}\hfill%
  \begin{subfigure}{.48\textwidth} 
    \centering
    \includegraphics[width=0.8\linewidth]{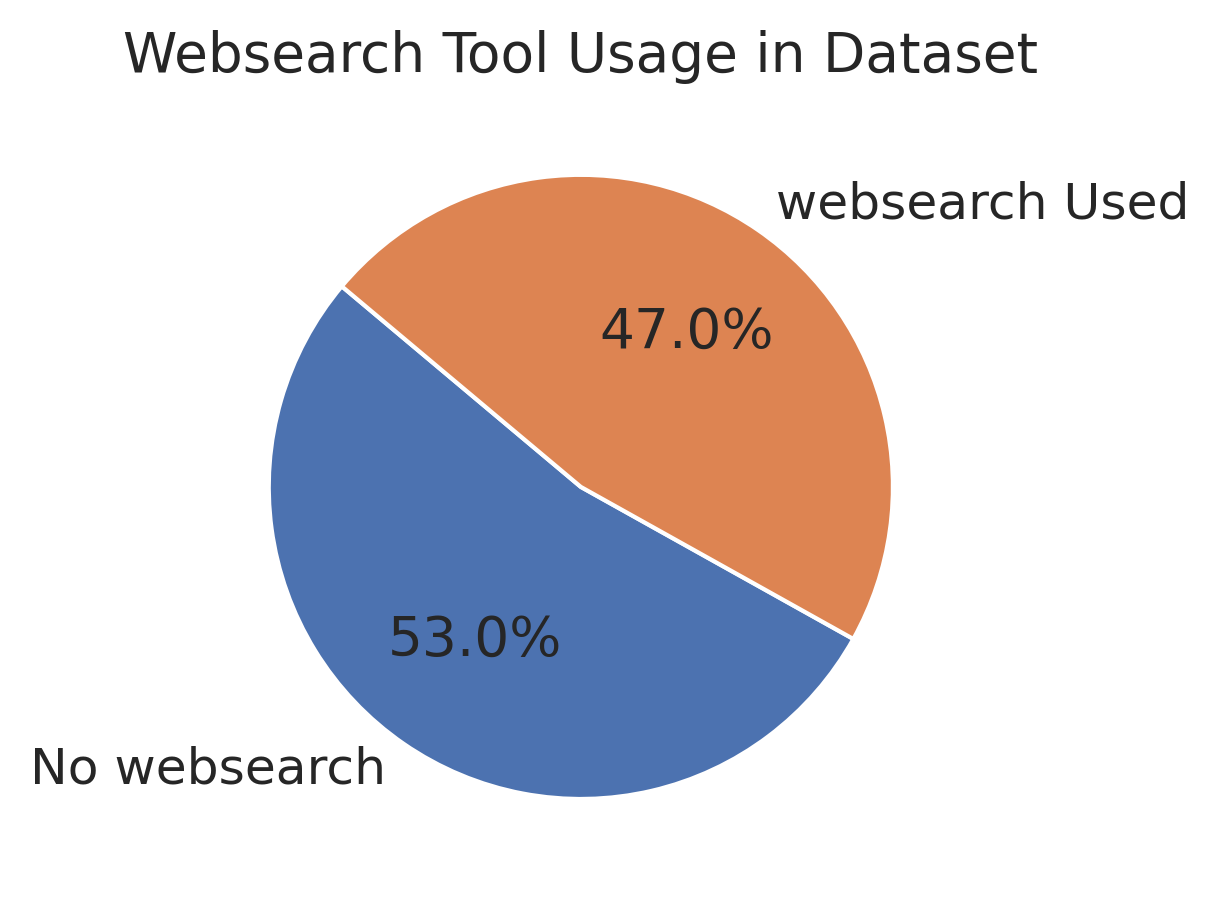}
    \caption{Overall distribution of Websearch Tool usage.}
    \label{fig:retrieval_usage_pie}
  \end{subfigure}
  
  % \caption{Visual exploration of the \texttt{FactualCS} dataset statistics, highlighting the stratified splits, category diversity, and strategic necessity of evidence retrieval.}
  \label{fig:dataset_stats}
\end{figure*}

We perform a comprehensive statistical analysis of \texttt{FactualCS} to characterize the distribution of hate speech types and the corresponding necessity for external evidence. The dataset consists of a total of 4,784 instances, which are partitioned into a training set of 3,912 samples, a validation set of 383 samples, and a test set of 489 samples, as detailed in Figure \ref{fig:split_sizes}. To prevent class imbalance during model training and evaluation, we employed a stratified splitting strategy, ensuring that the proportional representation of each hate speech category remains consistent across all three splits (Figure \ref{fig:hs_type_split_stacked}). In terms of class distribution, the dataset is semantically diverse: Stereotype Hate (SH) constitutes the largest portion at 32.8\%, followed by Non-factual Hate (NH) at 27.1\% and Dehumanization Hate (DH) at 23.5\%, with Misinformation (MH) and Conspiracy Hate (CH) accounting for 12.1\% and 4.4\% respectively (Figure \ref{fig:hs_type_pie}). A defining feature of our dataset is the strategic inclusion of reasoning traces for evidence retrieval. As shown in Figure \ref{fig:retrieval_usage_pie}, 47.0\% of the instances utilize the web search tool to ground the counterspeech. Crucially, this tool usage is strictly correlated with the hate category; as illustrated in Figure \ref{fig:hs_type_vs_retrieval}, factual categories (SH, MH, CH) predominantly trigger web searches to debunk claims, whereas abstract categories (DH, NH) rely on logical and moral refutation without requiring external data.

\begin{table}[t!]
    \centering
    \resizebox{.5\textwidth}{!}{
    \setlength{\tabcolsep}{1mm}
    \fontsize{9pt}{12pt}\selectfont
    \begin{tabular}{l|c|c|c|c}
        \hline
        \textbf{Metric} & \textbf{T-statistic} & \textbf{$p$-value} & \makecell{\textbf{Significant}} & \makecell{\textbf{Outperform}} \\
        \hline
        \multicolumn{5}{c}{\textbf{\texttt{FIRE} vs LLaMA-3.1-8B-Instruct (SFT)}} \\
        \hline
        ROUGE-1        & -9.41  & $2.05E{-19}$ & Yes & No \\
        ROUGE-2        & -9.75  & $1.28E{-20}$ & Yes & No \\
        ROUGE-L        & -11.19 & $5.63E{-26}$ & Yes & No \\
        BERTScore      & -11.33 & $1.48E{-26}$ & Yes & No \\
        METEOR         & -7.35  & $8.80E{-13}$ & Yes & No \\
        RepetitionRate & 0.35   & 0.727        & No  & Yes \\
        CoSIM          & -4.75  & $2.74E{-06}$ & Yes & No \\
        Toxicity       & 0.36   & 0.717        & No  & Yes \\
        Novelty        & 12.72  & $3.84E{-32}$ & Yes & Yes \\
        Diversity      & -0.35  & 0.727        & No  & Yes \\
        CatAcc         & 6.84   & $1.10E{-11}$ & Yes & Yes \\
        FactualScore   & 14.92  & $2.10E{-34}$ & Yes & Yes \\
        \hline
        \multicolumn{5}{c}{\textbf{\texttt{FIRE} vs LLaMA-3.1-8B-Instruct (FS)}} \\
        \hline
        ROUGE-1        & 4.31   & $1.99E{-05}$ & Yes & Yes \\
        ROUGE-2        & 2.53   & 0.012        & Yes & Yes \\
        ROUGE-L        & 0.97   & 0.331        & No  & Yes \\
        BERTScore      & 3.14   & $1.80E{-03}$ & Yes & Yes \\
        METEOR         & 11.33  & $1.53E{-26}$ & Yes & Yes \\
        RepetitionRate & 11.78  & $2.57E{-28}$ & Yes & Yes \\
        CoSIM          & 2.91   & $3.90E{-03}$ & Yes & Yes \\
        Toxicity       & -1.45  & 0.149        & No  & Yes \\
        Novelty        & 6.45   & $2.69E{-10}$ & Yes & Yes \\
        Diversity      & 11.78  & $2.57E{-28}$ & Yes & Yes \\
        CatAcc         & 9.27   & $4.80E{-20}$ & Yes & Yes \\
        FactualScore   & 18.63  & $7.40E{-41}$ & Yes & Yes \\
        \hline
    \end{tabular}
    }
    \caption{Statistical comparison of \texttt{FIRE} against LLaMA-3.1-8B-Instruct under supervised fine-tuning (SFT) and few-shot (FS) settings using paired t-tests. Positive T-statistic values indicate better performance by FIRE. Statistical significance is determined at \(p < 0.05\).}
    \label{tab:statistical_comparison_fire_llama}
\end{table}

\subsection{Statistical Significance Testing}
\label{stat_test}
To determine whether the observed performance differences between \textsc{FIRE} and best performing baseline model \textsc{LLaMA-3.1-8B-Instruct} are statistically meaningful, we conduct paired two-tailed t-tests across all evaluation metrics. Tests are performed separately for the supervised fine-tuning (SFT) and few-shot (FS) settings, comparing model outputs on identical test instances. Statistical significance is assessed at a threshold of $p < 0.05$.

As shown in Table~\ref{tab:statistical_comparison_fire_llama}, the SFT comparison reveals that \textsc{FIRE} achieves statistically significant improvements in factuality-oriented and task-specific metrics, including Novelty ($t=12.72, p<10^{-31}$), Category Accuracy ($t=6.84, p<10^{-11}$), and FactualScore ($t=14.92, p<10^{-33}$). In contrast, \textsc{LLaMA-3.1-8B-Instruct} attains higher scores on surface-level overlap and semantic similarity metrics such as ROUGE and BERTScore. Differences in Repetition Rate and Toxicity are not statistically significant, indicating comparable fluency and safety characteristics under supervised fine-tuning.

In the FS setting, \textsc{FIRE} consistently and significantly outperforms the baseline across nearly all metrics. Notable gains are observed in both generation quality and factual effectiveness, including ROUGE-1, ROUGE-2, METEOR, BERTScore, Category Accuracy ($t=9.27, p<10^{-19}$), and FactualScore ($t=18.63, p<10^{-40}$). Improvements in Novelty, Diversity, and Repetition Rate further suggest that \textsc{FIRE} produces more informative and less repetitive responses in the few-shot regime. Differences in Toxicity are not statistically significant, indicating that performance gains do not come at the expense of increased harmfulness.

The statistical analysis confirms that the observed improvements are robust and unlikely to be due to random variation.

\subsection{\textcolor{black}{Computational Efficiency Analysis}}
\label{app:efficiency}

\textcolor{black}{
In this study, the claim of being "memory-efficient" refers to the FIRE framework's ability to achieve state-of-the-art results using compact models without requiring the hardware footprint of larger, monolithic LLMs. As detailed in Table \ref{tab:efficiency_comparison}, FIRE utilizes two 1.7B parameter models (Hate Speech Analyst and Counterspeech Generator). Because these agents operate in distinct, sequential stages, the system only needs to load one 1.7B model into memory at any given time. This architectural choice reduces the peak hardware requirement from approximately 16GB of VRAM (typical for 7B--8B parameter models) to roughly 4GB of VRAM, representing a 75\% reduction in peak memory consumption. 
}

\textcolor{black}{
Furthermore, while FIRE involves multiple agentic steps and external tool usage, the overall inference time remains highly competitive. The slight overhead in FIRE's end-to-end latency (2.5 -- 3.0 seconds) is primarily attributable to the web search API round-trip during the HEAL phase. Despite this network latency, the framework's runtime remains comparable to single-pass generations from 7B and 8B baselines, proving that multi-agent decomposition can yield superior factual performance and significant memory savings without introducing prohibitive latency bottlenecks.
}

\begin{table}[h]
\centering
\small
\setlength{\tabcolsep}{4pt}
\begin{tabular}{lccc}
\toprule
\textbf{Model} & \textbf{Params} & \textbf{Peak VRAM} & \textbf{Latency} \\
\midrule
FIRE (Sequential) & 2$\times$1.7B & $\sim$4GB & 2.5--3.0s \\
Llama-3.1 8B-Instruct & 8B & $\sim$16GB & 2.0--2.8s \\
Qwen-2.5 7B-Instruct & 7B & $\sim$16GB & 1.8--2.5s \\
Mistral 7B-Instruct & 7B & $\sim$16GB & 1.7--2.4s \\
\bottomrule
\end{tabular}
\caption{Computational efficiency profiling comparing peak memory requirements and inference latency between FIRE and single-pass baselines.}
\label{tab:efficiency_comparison}
\end{table}

\subsection{\textcolor{black}{Hate Speech Analyst: Purity Score and Error Cascading}}
\label{app:error_cascading}

\textcolor{black}{
To address potential concerns regarding error propagation, where an initial misclassification by the Hate Speech Analyst (HSA) cascades into an incorrect downstream response, we analyze the distribution of the HSA's Purity Score ($\mathcal{P}$). This score acts as a confidence proxy by quantifying the homogeneity of the $k$-nearest neighbors retrieved from memory. A predefined threshold ($\tau = 0.62$) dictates whether the agent accepts the prediction with high confidence or triggers an uncertainty-aware "think again" mechanism. Based on an evaluation of 482 test samples, the purity score successfully isolates reliable predictions from ambiguous ones. As shown in Table \ref{tab:purity_distribution}, $\mathcal{P}$ exceeds the threshold in 88.4\% of instances, corresponding to a high accuracy of 73.2\%. Conversely, in the 11.6\% of cases where $\mathcal{P} < 0.62$, accuracy drops to 48.2\%. This 25.0\% gap confirms that $\mathcal{P}$ is a highly effective, correlated signal for downstream error likelihood. 
}

\textcolor{black}{
Furthermore, the relationship between purity and accuracy follows a clear monotonic trend: accuracy increases steadily from 46.9\% in the 0.4--0.6 range to 76.0\% in the high-confidence 0.8--1.0 range (peaking at 82.7\% when $\mathcal{P}=1.0$). By successfully flagging the 11.6\% of instances that are highly ambiguous and forcing the agent to explicitly reason over the top-two predicted categories, the framework actively intercepts and mitigates cascading errors prior to the final generation phase.
}

\begin{table}[h]
\centering
\small
\setlength{\tabcolsep}{6pt}
\begin{tabular}{lcc}
\toprule
\textbf{Condition} & \textbf{Samples (\%)} & \textbf{Accuracy} \\
\midrule
High Confidence ($\mathcal{P}{\ge}0.62$) & 88.4\% (426) & 73.2\% \\
Ambiguous ($\mathcal{P}{<}0.62$) & 11.6\% (56) & 48.2\% \\
\midrule
\textbf{Overall} & \textbf{100\%} (482) & \textbf{70.3\%} \\
\bottomrule
\end{tabular}
\caption{Distribution and accuracy breakdown of HSA predictions based on the purity score threshold.}
\label{tab:purity_distribution}
\end{table}

\subsection{Promting Strategies}
\label{prompt_strategies}
\subsubsection{ZeroShot Prompt}
We used the following system prompt for the Zero-Shot setting of baselines:
\begin{quote}
\textbf{System Instruction}: You are a hate speech analyst. Analyze the hate text and identify the target group, category of hate (misinfo\_hate, nonfactual\_hate, conspiracy\_hate, stereotype\_hate, dehumanization\_hate), and give reasoning behind your decision in \texttt{analyst\_reasoning} field. Generate concise and factual search queries and evidence (a single evidence) that will help verify or fact-check the hate speech. Generate a concise, factual counter-speech (a single counterspeech text) in response to the given hate text using its target group, hate type, and reasoning. Use retrieved evidence when available to correct misinformation or challenge harmful narratives. You output ONLY valid JSON. No markdown, no explanation, no commentary. Only return JSON with fields: \texttt{target\_group}, \texttt{hs\_type}, \texttt{analyst\_reasoning}, \texttt{retrieval\_query}, \texttt{retrieved\_evidence}, \texttt{counterspeech}.

\textbf{User Input}: [HATE\_SPEECH]\\
Analyze the hate speech and return ONLY JSON. Do NOT add \texttt{```json} or explanations.
\end{quote}

\subsubsection{FewShot Prompt}
For the Few-Shot setting, we utilized the same system instruction as the Zero-Shot setting but provided $5$ in-context examples to guide the model's reasoning and formatting:

\begin{quote}
\textbf{System Instruction}: [Same as Zero-Shot]

\textbf{Examples}:

\begin{enumerate}
    \item \textbf{User}: ``Weak beta male, hope he turns into a tranny after this traumatic experience, haha''
    
    \textbf{Assistant}: \{\\
    ``target\_group'': ``transgender people'',\\
    ``hs\_type'': ``dehumanization\_hate'',\\
    ``analyst\_reasoning'': ``This statement expresses a desire for harm to befall someone and uses a derogatory term for transgender people. It aims to strip them of their humanity by associating them with weakness and trauma.'',\\
    ``retrieval\_query'': null,\\
    ``retrieved\_evidence'': null,\\
    ``counterspeech'': ``Wishing harm on someone and using derogatory terms against transgender people is unacceptable and harmful. Transgender individuals deserve respect and dignity, like everyone else.''\\
    \}

    \item \textbf{User}: ``The Democrat Party has ONE MISSION - to destroy the USA and if YOU press the Dummy lever - you are helping... LINK''
    
    \textbf{Assistant}: \{\\
    ``target\_group'': ``Democrat Party'',\\
    ``hs\_type'': ``misinfo\_hate'',\\
    ``analyst\_reasoning'': ``This statement makes a false claim about the Democrat Party's mission. It needs fact-checking to determine accuracy.'',\\
    ``retrieval\_query'': ``Democrat Party mission statement'',\\
    ``retrieved\_evidence'': ``The Democratic Party states its mission is to support working families, equal opportunity, and a healthy future for all Americans.'',\\
    ``counterspeech'': ``The Democratic Party's mission focuses on supporting working families and ensuring equal opportunity, not destroying the USA.''\\
    \}

     \item \textbf{User}: ``Women are complaining because they cant get highpaying jobs. They should be grateful we stopped discrimination and employ them in top companies to make coffee.''
    
    \textbf{Assistant}: \{\\
    ``target\_group'': ``women'',\\
    ``hs\_type'': ``stereotype\_hate'',\\
    ``analyst\_reasoning'': ``This statement relies on the stereotype that women are only suitable for menial jobs and that they are ungrateful despite progress in gender equality. It diminishes their capabilities and aspirations.'',\\
    ``retrieval\_query'': ``gender pay gap statistics'',\\
    ``retrieved\_evidence'': ``In 2022, women earned 84 cents for every dollar earned by men among full-time workers.'',\\
    ``counterspeech'': ``Women have significantly contributed to various fields throughout history; Marie Curie, for example, pioneered radioactivity research. The gender pay gap persists, but women's ambitions go far beyond stereotypical roles; they aim to lead and innovate, deserving equal opportunities and recognition.''\\
    \}

    \item \textbf{User}: ``If women dont stop complaining, then men will find a solution. If a woman does not submit, then there will be war.''
    
    \textbf{Assistant}: \{\\
    ``target\_group'': ``women'',\\
    ``hs\_type'': ``nonfactual\_hate'',\\
    ``analyst\_reasoning'': ``This statement promotes male dominance and threatens violence against women who do not conform to traditional gender roles. It does not contain factual information but expresses a harmful opinion.'',\\
    ``retrieval\_query'': null,\\
    ``retrieved\_evidence'': null,\\
    ``counterspeech'': ``I understand your frustration; it's tough when people don't see eye-to-eye. Let's strive for mutual respect and open communication in our interactions.''\\
    \}

     \item \textbf{User}: ``We are ruled by a small Jewish elite that exploit the weak.''
    
    \textbf{Assistant}: \{\\
    ``target\_group'': ``Jewish people'',\\
    ``hs\_type'': ``conspiracy\_hate'',\\
    ``analyst\_reasoning'': ``This statement promotes the antisemitic conspiracy theory that Jewish people secretly control governments and exploit others for their benefit.'',\\
    ``retrieval\_query'': ``Jewish people controlling the world'',\\
    ``retrieved\_evidence'': ``The claim that Jewish people secretly control the world is a long-running antisemitic conspiracy theory.'',\\
    ``counterspeech'': ``The claim that a Jewish elite rules and exploits is a conspiracy theory. This antisemitic trope is rooted in fabricated texts like 'The Protocols of the Elders of Zion'. These lies have been repeatedly discredited. Jews account for approximately 0.2 percent of the global population.''\\
    \}

\end{enumerate}

\textbf{User Input}: ``[HATE\_SPEECH]"\\
Analyze the hate speech and return ONLY JSON. Do NOT add \texttt{```json} or explanations.
\end{quote}

\subsubsection{Retrival-Based Prompt}
In the Retrieval-Based setting, we dynamically select the most similar example from the training set using cosine similarity on sentence embeddings (all-mpnet-base-v2). This retrieved exemplar is prepended to the prompt context.

\begin{quote}
\textbf{System Instruction}: [Same as Zero-Shot]

\textbf{Retrieved Exemplar (Top-1)}: ``[RETRIEVED\_HS\_EXAMPLE]''
\textbf{Assistant}: \{ ``target\_group'': ``...'', ``hs\_type'': ``...'', ... \}

\textbf{User Input}: ``[HATE\_SPEECH]"\\
Analyze the hate speech and return ONLY JSON.
\end{quote}

\subsection{Computing Information}
Our research is conducted on the NVIDIA RTX A100 with $80$ GB RAM GPU.

\subsection{Hyper-parameter Information}
\begin{itemize}
    \item Precision: BF16
    \item Temperature: 0.0 (Greedy decoding)
    \item Baselines Batch Size: 16
    \item Baselines SFT Learning rate: 2e-4
    \item Max New Tokens: 512
    \item Retrieval Encoder: sentence-transformers/all-mpnet-base-v2
\end{itemize}

\subsection{\textcolor{black}{Manual Audit of Websearch Tool Quality}}
\label{app:web_search_audit}

\textcolor{black}{
To directly address concerns regarding the reliability and potential retrieval of harmful or fake web evidence, we conducted a manual audit on a randomly sampled subset of 100 evidence instances retrieved during the \texttt{HEAL} phase. We categorized the sources into four tiers: Institutional/Academic, Mainstream Informational, Social-Media-Style, and Potentially Unreliable.
}

\textcolor{black}{
As detailed in Table \ref{tab:search_audit}, the combination of the DuckDuckGo safe-search API and our framework's strict filtering effectively suppresses low-quality content. The vast majority of retrieved evidence (92\%) originates from reliable domains, with 58\% coming from institutional or academic sources (e.g., WHO, PubMed/NCBI, ScienceDirect) and 34\% from mainstream informational websites (e.g., Wikipedia). Only 6\% originated from social-media-style platforms (e.g., Reddit), and a mere 2\% were judged as potentially low-quality or weakly reliable. This low incidence of content from malicious sources confirms that \texttt{FIRE}'s retrieval mechanism acts as a robust factual safeguard rather than an amplifier of misinformation. Furthermore, as a secondary defense against the 2\% of potentially unreliable sources that bypass the safe-search filter, the Counterspeech Generator's system prompt explicitly instructs the agent to critically evaluate retrieved text and strictly prohibits the propagation of any hate speech claims.
}

\begin{table}[h]
\centering
\small
\setlength{\tabcolsep}{6pt}
\begin{tabular}{lc}
\toprule
\textbf{Source Category} & \textbf{Retrieval Frequency} \\
\midrule
Institutional / Academic & 58\% \\
Mainstream Informational & 34\% \\
Social-Media-Style & 6\% \\
Potentially Low-Quality & 2\% \\
\bottomrule
\end{tabular}
\caption{Manual audit of 100 retrieved web search instances categorized by source reliability.}
\label{tab:search_audit}
\end{table}

\subsection{\textcolor{black}{Human Evaluation Protocol}}
\label{app:human_eval_rubric}

\textcolor{black}{
To ensure rigorous subjective evaluation, we employed 30 expert annotators who were provided with formal definitions of hate speech and counterspeech. The annotators were tasked with a comparative ranking of model outputs from 1 (Best) to 4 (Worst) across four distinct dimensions: Independent Counterspeech (ICS), Adequacy/Fluency (Ad), Contextual Relevance (CoRl), and Argumentative Effectiveness (ArgE). Following the ranking phase, responses were categorized into Best, Above Average, Average, and Worst. To calibrate the annotators and ensure consistency, we provided explicit anchor examples prior to the task. For instance, a "Good/Best" response for Argumentative Effectiveness directly refutes the underlying propositional claim using logical evidence without resorting to counter-attacks. Conversely, a "Bad/Worst" response either hallucinates facts, utilizes toxic language, or completely misidentifies the target group's protected characteristic. This structured calibration ensured high reliability in the final comparative rankings presented in Section \ref{sec:human_evaluation_section}. The exact instructions provided to annotators were as follows:}

\vspace{1mm}
\noindent\textbf{\textcolor{black}{Core Definitions}}\\
\textcolor{black}{
\textbf{Hate Speech (HS):} Communication that attacks or uses pejorative/discriminatory language against a person or group based on identity factors (religion, race, descent, gender, etc.).\\
\textbf{Counterspeech (CS):} A tactic to counter hate speech or misinformation by presenting an alternative narrative, using empathy, and challenging hate narratives without censorship or reciprocal hate.\\
\textit{Anchor Example (HS):} ``All Muslims are terrorists.'' \\
\textit{Anchor Example (CS):} ``This statement is a gross generalization that is not supported by facts. The vast majority of Muslims are peaceful...''
}

\vspace{1mm}
\noindent\textbf{\textcolor{black}{Evaluation Metrics (Ranked 1 [Best] to 4 [Worst])}}
\textcolor{black}{
\begin{itemize}[leftmargin=*]
    \item \textbf{Independent CS (IC):} Measures if the CS stands alone meaningfully without context. \\ \textit{Bad:} ``No, they're not.'' \quad \textit{Good:} ``This statement is a gross generalization not supported by facts.''
    \item \textbf{Fluency (F):} Measures grammatical soundness and coherence. \\ \textit{Bad:} ``Muslims are good people.'' \quad \textit{Good:} ``The vast majority of Muslims are peaceful and law-abiding citizens.''
    \item \textbf{Contextual Relevance (CR):} Addresses central/implicit aspects of the HS (Subject, False Claim, Stereotype). \\ \textit{Bad:} ``You're wrong.'' \quad \textit{Good:} ``It is unfair and unjust to label an entire group of people based on the actions of a few extremists.''
    \item \textbf{Argumentative Effectiveness (AE):} Provides a coherent, persuasive, logically sound argument. \\ \textit{Bad:} ``No, they're not.'' \quad \textit{Good:} ``This statement is a gross generalization... We should focus on building bridges between different communities.''
\end{itemize}
}

\vspace{1mm}
\noindent\textbf{\textcolor{black}{Example Annotation Interface}}\\
\textcolor{black}{
For each section, annotators were given an HS statement and four CS generations (A, B, C, D) and asked to rank them for each of the four metrics. A real task instance is shown below:
}

\begin{quote}
\small
\textbf{\textcolor{black}{HS:}} \textcolor{black}{Muslim are invading our country.}

\textbf{\textcolor{black}{CS-A:}} \textcolor{black}{Muslims enrich our country with their diverse culture, traditions, and values. Their presence brings beauty and unity, making our nation stronger and more vibrant...}

\textbf{\textcolor{black}{CS-B:}} \textcolor{black}{Muslims are contributing wonderfully to our society, spreading a culture full of rich history, unique traditions, and incredible cuisine that brings joy...}

\textbf{\textcolor{black}{CS-C:}} \textcolor{black}{The claim that Muslims are invading our country is false and it spreads misinformation about immigration and demographic changes. It stokes fear and hatred by suggesting that a specific group is deliberately undermining the native population. But the reality is shaped by complex social and economic factors, not hostile intent...}

\textbf{\textcolor{black}{CS-D:}} \textcolor{black}{I am a muslim and i am not invading your country. i am a muslim and i am not invading your country. i am a muslim and i am not invading your country... }
\end{quote}

\end{document}